\PassOptionsToPackage{table,dvipsnames}{xcolor}
\documentclass[]{opendatalab}
\usepackage{amssymb}
\usepackage{float}
\usepackage{longtable}
\usepackage{listings}
\usepackage{makecell}
\usepackage{xspace}
\definecolor{codegreen}{rgb}{0,0.6,0}
\definecolor{codegray}{rgb}{0.5,0.5,0.5}
\definecolor{codepurple}{rgb}{0.58,0,0.82}
\definecolor{backcolour}{rgb}{0.95,0.95,0.92}
\definecolor{promptcolor}{HTML}{D1D0F2}
\definecolor{promptcolorheader}{HTML}{bdbcec}
\newcommand{\mineru}{MinerU-Diffusion\xspace}
\newcommand{\promptbox}[2]{
\begin{tcolorbox}[
top=0.3em,bottom=0.3em,left=0.5em,right=0.5em,
toptitle=0.3em,bottomtitle=0.2em,boxsep=0pt,
colframe=promptcolorheader,colback=promptcolor!50,boxrule=0.5pt,
]
\footnotesize
\end{tcolorbox}
}
\lstdefinestyle{mystyle}{
    backgroundcolor=\color{backcolour},   
    commentstyle=\color{codegreen},
    keywordstyle=\color{magenta},
    numberstyle=\tiny\color{codegray},
    stringstyle=\color{codepurple},
    basicstyle=\ttfamily\footnotesize,
    breakatwhitespace=false,         
    breaklines=true,                 
    captionpos=b,                    
    keepspaces=true,                 
    numbers=left,                    
    numbersep=5pt,                  
    showspaces=false,                
    showstringspaces=false,
    showtabs=false,                  
    tabsize=2
}
\makeatletter
\usepackage{marvosym}
\title{MinerU-Diffusion: Rethinking Document OCR as Inverse Rendering via Diffusion Decoding}

\author[1*]{Hejun Dong}
\author[2*]{Junbo Niu}
\author[1\textrm{\Letter} \ddagger]{Bin Wang}
\author[2]{Weijun Zeng}
\author[2\textrm{\Letter}]{Wentao Zhang}
\author[1\textrm{\Letter}]{Conghui He}

\affiliation[1]{Shanghai Artificial Intelligence Laboratory, OpenDataLab}
\affiliation[2]{Peking University}

\abstract{
Optical character recognition (OCR) has evolved from line-level transcription to structured document parsing, requiring models to recover long-form sequences containing layout, tables, and formulas. Despite recent advances in vision-language models, most existing systems rely on autoregressive decoding, which introduces sequential latency and amplifies error propagation in long documents. In this work, we revisit document OCR from an inverse rendering perspective, arguing that left-to-right causal generation is an artifact of serialization rather than an intrinsic property of the task. Motivated by this insight, we propose \textbf{MinerU-Diffusion}, a unified diffusion-based framework that replaces autoregressive sequential decoding with parallel diffusion denoising under visual conditioning. MinerU-Diffusion employs a block-wise diffusion decoder and an uncertainty-driven curriculum learning strategy to enable stable training and efficient long-sequence inference. Extensive experiments demonstrate that MinerU-Diffusion consistently improves robustness while achieving up to 3.2× faster decoding compared to autoregressive baselines. Evaluations on the proposed Semantic Shuffle benchmark further confirm its reduced dependence on linguistic priors and stronger visual OCR capability.
}

\metadata[* Equal contribution\quad $\textrm{\Letter}$ Corresponding author \quad $\ddagger$ Project leader]{}
\correspondence{Conghui He, \email{heconghui@pjlab.org.cn}}
\metadata[Code]{\url{https://github.com/opendatalab/MinerU-Diffusion}}
\metadata[Model]{\url{https://huggingface.co/opendatalab/MinerU-Diffusion-V1-0320-2.5B}}
\date{\today}

\begin{document}

\maketitle
\newpage
\tableofcontents
\newpage

\begin{figure}[H]
\centering
\begin{minipage}[t]{\linewidth}
\centering
\includegraphics[width=\linewidth]{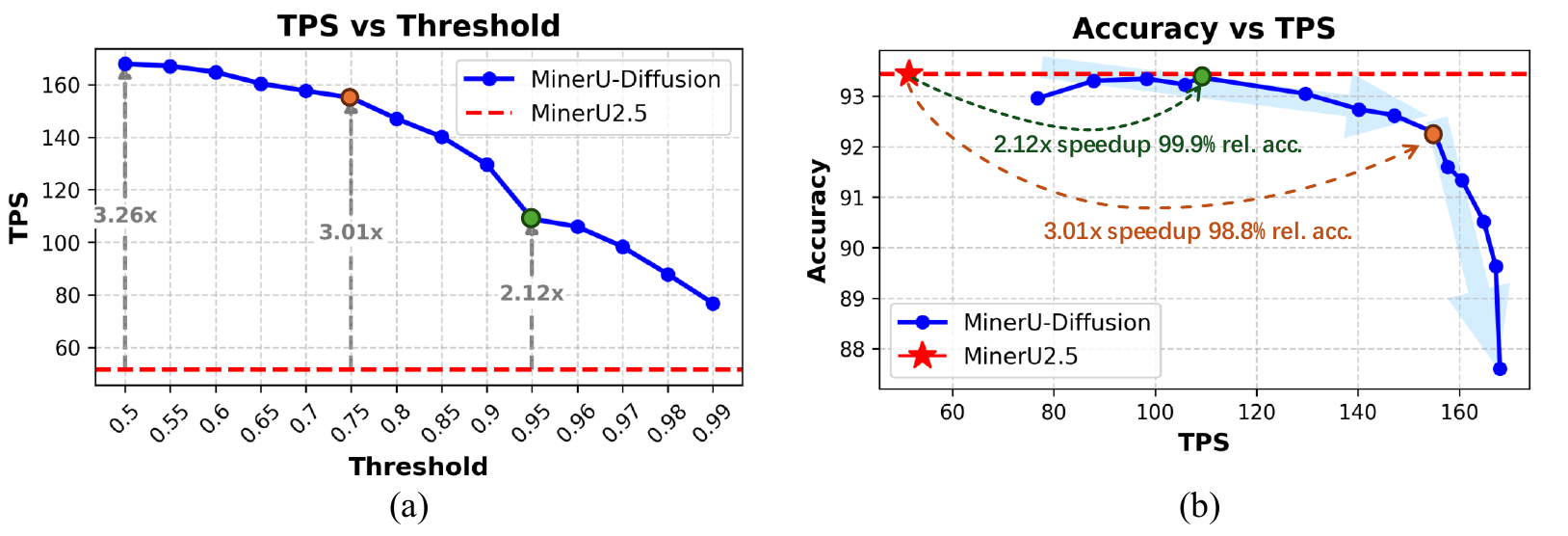}
\end{minipage}

\medskip

\begin{minipage}[t]{\linewidth}
\centering
\includegraphics[width=\linewidth]{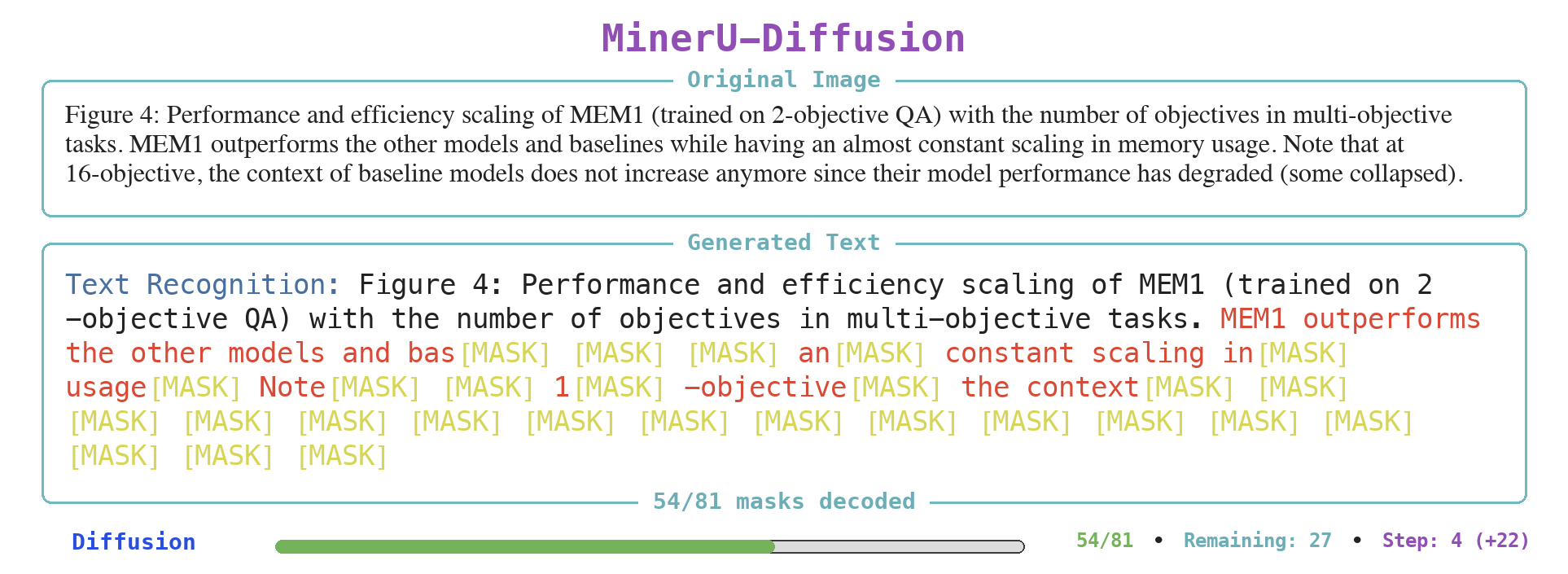}

\smallskip

(c)
\end{minipage}

\caption{\textbf{(a)} The confidence threshold controls decoding parallelism in MinerU-Diffusion. Compared to MinerU2.5, our method achieves up to 3.26× speedup. \textbf{(b)} MinerU-Diffusion maintains a strong accuracy–efficiency trade-off, achieving 2.12× speedup with 99.9\% and 3.01× speedup with 98.8\% relative accuracy. \textbf{(c)} Diffusion decoding progressively reconstructs structured text from masked tokens under visual conditioning: black tokens are confirmed, red tokens are being updated, and yellow tokens remain masked, enabling parallel generation with global consistency, in contrast to autoregressive left-to-right decoding.}
\label{fig:front_overview}
\end{figure}

\section{Introduction}
\label{sec:intro}

In recent years, Vision-Language Models (VLMs) \cite{achiam2023gpt, comanici2025gemini, wang2024qwen2, zhu2025internvl3, guo2025seed1, li2024llava, zhang2024internlm, zeng2026research, niu2025ovo} have become the dominant paradigm for document Optical Character Recognition (OCR) \cite{niu2025mineru2, li2025monkeyocr, li2025dots, wei2025deepseek, cui2025paddleocr}. These models encode textual images into visual representations and generate structured text through left-to-right autoregressive decoding \cite{vaswani2017attention}, achieving strong performance across benchmarks \cite{ouyang2025omnidocbench,fu2024ocrbench, wang2024unimernet}. Despite architectural unification and scaling, the decoding process remains strictly sequential. This design introduces efficiency and reliability bottlenecks when parsing long documents and complex layouts, particularly in highly structured scenarios such as tables and formulas \cite{ouyang2025omnidocbench}.

From the perspective of task formulation, a high-quality OCR system should primarily depend on authentic visual evidence to perform character-level recognition, rather than relying on semantic completion from a language model. However, autoregressive formulations implicitly cast OCR as language-conditioned reconstruction, where textual outputs are generated under strong linguistic priors. When visual signals are weak or semantic constraints are disrupted, models tend to over-rely on these priors, leading to semantic hallucinations and cumulative errors. Experiments on the Semantic Shuffle benchmark confirm that disrupting semantic structure causes substantial performance degradation in AR-based OCR systems. This fragility stems not merely from data or training strategy, but from the causal factorization inherent in autoregressive decoding.

In contrast, Diffusion Language Models (DLMs) \cite{labs2025mercury, google_deepmind_gemini_diffusion_2025, song2025seed,nie2025large, dream2025}, based on discrete diffusion processes, provide a modeling paradigm better aligned with the structural characteristics of document OCR tasks. Masked diffusion models assume conditional independence among tokens given partially observed sequences and visual inputs \cite{you2025lladav}, which is reasonable in OCR scenarios where the mapping between image content and text is largely deterministic with limited semantic ambiguity. This property allows models to exploit the local consistency of visual signals and perform parallel decoding of long textual segments while maintaining global coherence and accuracy. Compared with autoregressive generation for open-ended text tasks \cite{liu2023visual}, diffusion decoding naturally matches the deterministic nature of document OCR \cite{ouyang2025omnidocbench}. Moreover, DLMs support parallel multi-token updates, significantly improving inference efficiency for long-document parsing. Although current diffusion-based VLMs \cite{cheng2025sdar-vl, you2025lladav, liu2025llavida, yang2025mmada} still encounter issues such as instability in long sequences, repetition, and hallucination in high-resolution settings, these limitations can be progressively alleviated through improved model design and training strategies, making masked diffusion a promising and principled alternative for accurate and efficient document OCR modeling.

Motivated by these observations, we formulate document OCR explicitly as an inverse rendering problem under visual conditioning, shifting from autoregressive causal decoding to diffusion-based decoding. We propose MinerU-Diffusion, a unified diffusion-based parsing framework tailored for document OCR. Centered on block-wise diffusion decoding~\cite{arriola2025block, cheng2025sdar, wang2025revolutionizing} and coupled with an uncertainty-driven curriculum learning strategy, MinerU-Diffusion enables global parallel reconstruction for document OCR. While maintaining high recognition accuracy, MinerU-Diffusion significantly improves long-sequence inference efficiency and effectively mitigates semantic hallucination and cumulative error propagation observed in autoregressive decoding. Extensive experiments demonstrate that MinerU-Diffusion achieves performance on par with state-of-the-art approaches across multiple challenging document parsing benchmarks and semantic perturbation settings, while attaining a more favorable balance among recognition accuracy, robustness, and decoding efficiency. More examples are provided in \hyperref[app:examples]{Appendix~\ref*{app:examples}}.

\section{Related Works}

\noindent\textbf{Vision-Language Models for Document OCR.}
Driven by large-scale pre-training, document OCR has evolved from traditional modular pipelines \cite{cui2025paddleocr,livathinos2025docling,vik2024marker,wang2024mineru} toward end-to-end Vision-Language Models (VLMs) that generate structured text directly from pixels \cite{li2025monkeyocr, li2025dots, poznanski2025olmocr, nanonets2025, comanici2025gemini, achiam2023gpt, guo2025seed1, wei2025deepseek}. Representative systems such as MinerU2.5 \cite{niu2025mineru2} and PaddleOCR-VL \cite{cui2025paddleocr} formulate document OCR as sequence generation and rely on autoregressive (AR) decoders to produce text token by token. While this unified paradigm simplifies traditional pipelines and improves cross-domain generalization, it inherits structural limitations from causal left-to-right decoding. Inference latency scales linearly with output length, limiting efficiency in long-document scenarios. Moreover, the strong coupling between generation order and linguistic context encourages reliance on language priors, which may compromise robustness when visual evidence is ambiguous or semantic structure is disrupted. These limitations motivate alternative decoding paradigms that enable global dependency modeling and reduce dependence on unidirectional factorization.

\noindent\textbf{Masked Diffusion Language Models.}
Diffusion Language Models (DLMs) \cite{labs2025mercury, google_deepmind_gemini_diffusion_2025, song2025seed,nie2025large, dream2025} offer a non-autoregressive generative framework based on discrete diffusion processes. In masked diffusion, tokens in a clean sequence \(x_0\) are progressively replaced by mask tokens \([{\rm MASK}]\) under a continuous corruption schedule \(t \in [0, 1]\), yielding a noised sequence \(x_t\). The forward process is defined as:
\begin{align}
\label{process}
q(x_t \mid x_0) = \prod_{i=1}^{n} \mathrm{Cat}\!\left(x_t^i; (1 - t)\delta_{x_0^i} + t \delta_{\text{[MASK]}}\right).
\end{align}

The corresponding training objective can be derived from maximum likelihood estimation \cite{shi2024simplified, sahoo2024simple, zheng2024masked, ou2024your}, resulting in an evidence lower bound (ELBO) on $\log p_{\theta}(x)$:
\begin{align}
\label{dlmobj}
\mathcal{J}_{\text{full}}(x_0, Q, \theta) 
= \int_{0}^{1} \frac{1}{t |x_0|} 
\mathbb{E}_{q(x_t \mid x_0)} 
\left[ 
\sum_{i: x_t^i = \text{[MASK]}} 
\log p_{\theta}(x_0^i \mid x_t, Q) 
\right] dt,
\end{align}
where $Q$ denotes the prompt and $|x_0|$ is the sequence length.

To enhance scalability, block diffusion models \cite{arriola2025block, JetAstra2025, wu2025fastv2, wang2025revolutionizing} introduce block-wise attention mechanisms that balance the optimization stability of autoregressive (AR) training with the parallel sampling efficiency of diffusion-based generation. Their structured attention patterns naturally enable KV-cache reuse, alleviating the inference latency commonly observed in full-attention DLMs \cite{hu2025accelerating, liu2025dllm, ma2025dkv, arriola2025block, yu2025dimple, wu2025fast}.

Masked diffusion models are structurally well aligned with the characteristics of document OCR tasks. In document OCR, the target text typically exhibits a near-deterministic mapping to the textual content present in the image, with limited semantic ambiguity. Under this setting, the conditional independence assumption underlying masked diffusion—that each token can be predicted independently given the input and partially observed sequence, as illustrated in~\Cref{fig:render}—becomes considerably more reasonable than in open-ended language generation \cite{man2026dodo, zhu2025llada, nie2025large}. This alignment allows the model to decode long-range text spans in parallel without sacrificing consistency.

Therefore, diffusion-based decoding is not merely an efficiency-oriented alternative to autoregressive methods. Instead, it constitutes a modeling paradigm that is inherently better matched to the structural properties of OCR, offering both theoretical justification and practical advantages for large-scale text recognition.

\section{Method}
\subsection{Problem Formulation: Inverse Rendering via Diffusion}
We model document OCR \cite{ouyang2025omnidocbench} as the inverse rendering of a unified structured token sequence:
\begin{equation}
y = \left(y^{(1)}, \dots, y^{(L)}\right) \in \mathcal{V}^L,
\end{equation}

\begin{figure}[!t]
    \centering
    \includegraphics[width=\linewidth]{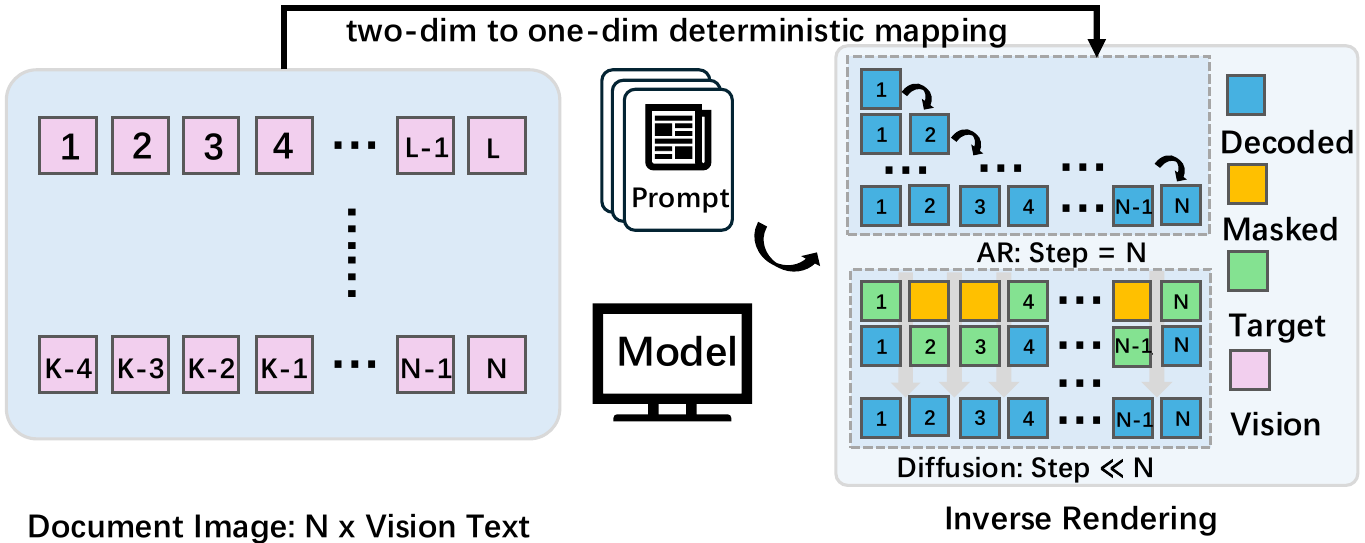}
    \caption{Overview of the document OCR inverse rendering process via different decoding methods.. The model maps a 2D document image to a 1D token sequence for decoding through autoregressive and diffusion-based methods.}
    \label{fig:render}
\end{figure}

where $\mathcal{V}$ is a shared vocabulary encompassing text symbols, layout markers, table delimiters, and mathematical operators. This unified representation enables the encoding of heterogeneous document elements—such as paragraphs, tables, formulas, and reading order—within a single sequential interface.

Although serialized as a one-dimensional sequence, $y$ corresponds to an underlying two-dimensional document structure. The statistical dependencies between tokens arise primarily from spatial arrangement, layout regularities, and formatting constraints, rather than from an intrinsic causal generation order. Therefore, the serialization order should be viewed as an implementation artifact introduced for representation convenience, rather than a fundamental property of the document generation process. In this sense, OCR output is more naturally modeled as a spatially coupled discrete random field, rather than a strictly directional sequence.

Document OCR can be framed as posterior inference over a latent structured token sequence, where the input document \(x\) serves as partial and noisy evidence, constraining both token identities and spatial positioning. Traditional OCR systems typically parameterize the posterior through autoregressive decompositions \cite{niu2025mineru2, cui2025paddleocr, wei2025deepseek, li2025dots}, which impose a fixed causal order and limit the ability to model document structure globally. In contrast, as illustrated in~\Cref{fig:render}, diffusion-based decoding methods \cite{arriola2025block, nie2025large} introduce a discrete diffusion process that avoids a fixed causal ordering, enabling global iterative refinement under visual conditioning, which naturally aligns with the structural properties of OCR tasks. The conditional independence assumption inherent in masked diffusion models \cite{nie2025large} —i.e., each token can be independently predicted given the input and partially observed sequence—becomes particularly reasonable in OCR, where the target text has a one-to-one correspondence with the text in the image. Through multiple denoising iterations, diffusion models jointly update all tokens across the entire sequence, circumventing the single-pass update limitation of autoregressive decoding, and providing a more structurally aligned approximation to the posterior distribution \( p(y \mid x) \).

\subsection{\mineru: Unified Diffusion Architecture for OCR}

A straightforward implementation of discrete diffusion for OCR is to apply a full-attention dLM \cite{nie2025large, dream2025, yu2025dimple} over the entire token sequence at each denoising step. However, when scaling to long structured documents, such a design suffers from fundamental structural and computational limitations \cite{wu2025fast}.

Full self-attention incurs quadratic complexity $O(L^2)$ with respect to sequence length $L$, making it computationally expensive for long structured documents with thousands of tokens. Moreover, full-attention diffusion operates globally, which introduces positional instability as early denoising errors can propagate across the sequence. Unlike autoregressive decoding \cite{achiam2023gpt}, global diffusion lacks structural anchoring and is prone to long-range drift. Additionally, document structures exhibit strong locality, with high intra-region consistency and weak long-range dependencies. Full-attention unnecessarily couples independent regions, conflicting with the spatially constrained posterior structure of document OCR.

\begin{figure}[!t]
    \centering
    \includegraphics[width=\linewidth]{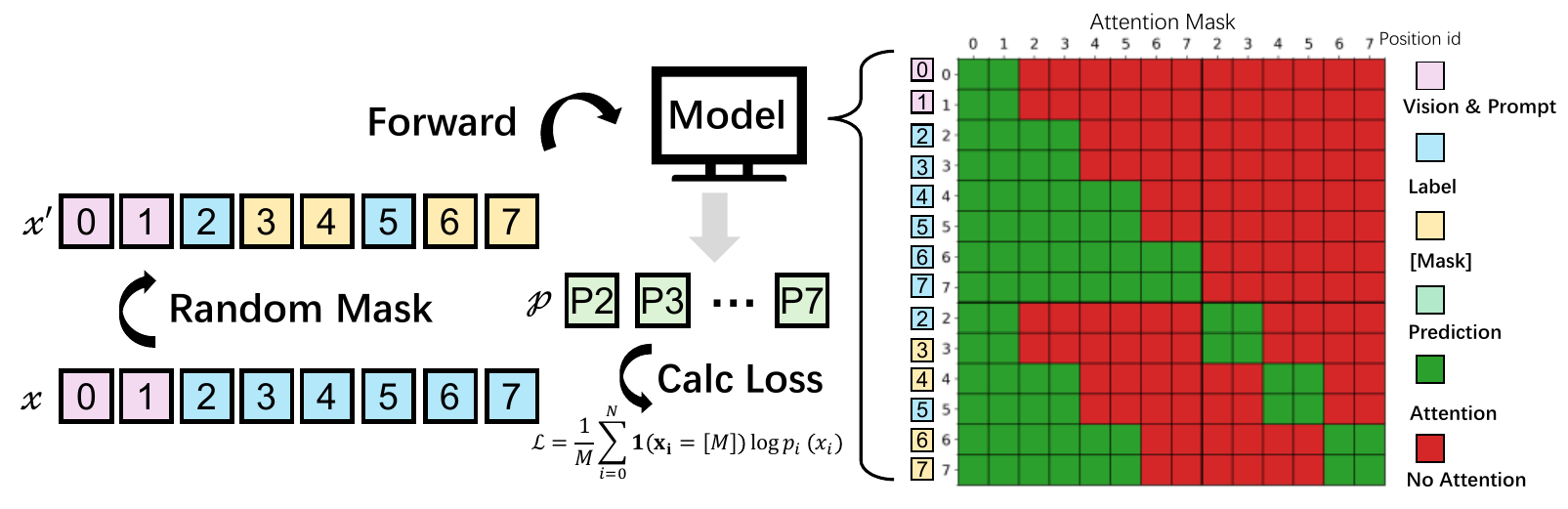}
    \caption{Training of MinerU-Diffusion. \textbf{Left:} the target token sequence is randomly masked to form a partially observed input, and the model predicts only the masked positions under visual and prompt conditioning. \textbf{Right:} the structured block-attention mask used during training, where tokens attend bidirectionally within each block and causally to all preceding blocks, enabling parallel diffusion refinement within blocks while preserving coarse autoregressive structure across blocks.}
    \label{fig:train}
\end{figure}

These observations suggest that purely full-attention diffusion is not structurally aligned with OCR. To address these limitations, we introduce \mineru, a block-attention \cite{arriola2025block, cheng2025sdar} dVLM that incorporates structural locality into posterior refinement, as illustrated in \Cref{fig:train}. The output sequence is partitioned into $B$ contiguous blocks:
\begin{equation}
y = (y^{(1)}, \dots, y^{(B)}), 
\quad y^{(b)} \in \mathcal{V}^{L'}, 
\quad L = B L'.
\end{equation}

Rather than modeling the entire sequence as a monolithic denoising problem, we factorize the conditional posterior:
\begin{equation}
p_\theta(y \mid x) = \prod_{b=1}^{B} p_\theta\big(y^{(b)} \mid y^{(<b)}, x\big),
\end{equation}
where $y^{(<b)}$ denotes all preceding blocks. Within each block, diffusion operates locally:
\begin{equation}
p_\theta\big(y^{(b)}_{t-1} \mid y^{(b)}_t, y^{(<b)}, x\big).
\end{equation}

This hybrid factorization introduces coarse-grained autoregressive structure across blocks and parallel diffusion refinement within blocks. Block boundaries serve as structural anchors, preventing long-range alignment drift, while preserving parallel efficiency inside each block.

At each denoising step, a structured attention mask is applied. Tokens can attend fully to tokens within the same block, causally to tokens in preceding blocks, and not to tokens in future blocks, as shown in \Cref{fig:train}. Formally, the attention mask $\mathcal{M}_{ij}$ is defined as:
\begin{equation}
\mathcal{M}_{ij}
=
\begin{cases}
1, & \text{if } b(i) = b(j), \\
1, & \text{if } b(j) < b(i), \\
0, & \text{otherwise}.
\end{cases}
\end{equation}
where $b(i)$ denotes the block index of token $i$. This structured masking reduces unnecessary global coupling, stabilizes positional alignment, and ensures that decoding errors remain locally bounded.

\mineru conditions the diffusion process on native-scale visual features \cite{niu2025mineru2, niu2025native, wang2024qwen2, dehghani2023patch}, ensuring that posterior refinement remains grounded in visual evidence. Compared to full-attention diffusion, block-attention reduces complexity from $O(L^2)$ to $O(B L'^2)$. The causal structure across blocks enables efficient KV-caching during inference, while maintaining parallel decoding within each block. Overall, \mineru provides a structurally grounded and computationally scalable diffusion architecture tailored for document OCR.

\subsection{Two-Stage Curriculum Learning with Uncertainty-Driven Refinement}
\label{sec:two-stage-curriculum}
To fully leverage large-scale heterogeneous data and alleviate the performance bottleneck caused by noisy labels, we propose a two-stage curriculum learning framework to train the \mineru. Compared with autoregressive (AR) models that generate tokens in a fixed order, diffusion models can decode tokens in any order, introducing more intricate inter-token dependencies that make training less stable and more sensitive to noise—often requiring larger datasets and more carefully tuned training strategies \cite{song2025seed, ni2025diffusion}. Furthermore, random masking modeling typically achieves lower data utilization efficiency compared to AR models at the same data scale. This is because AR models make predictions based on the complete prefix information at each step, while random masking disperses the supervisory signal, reducing the density of effective conditional information. To address this, we divide the dataset $\mathcal{D}$ into two subsets: $\mathcal{D}_{\text{base}}$, which is easier to train, and $\mathcal{D}_{\text{hard}}$, which is more challenging. We first train on $\mathcal{D}_{\text{base}}$ to establish the foundational structure understanding, then fine-tune on $\mathcal{D}_{\text{hard}}$ to improve the model's robustness to noisy labels and boundary precision. This framework effectively resolves the optimization complexity and low data utilization efficiency faced by diffusion models in any-order modeling.


\subsubsection{Stage I: Diversity-Driven Foundational Learning}
In the first stage, our goal is to establish robust foundational representations and general parsing abilities across multiple document understanding tasks. To this end, we construct a large-scale, diverse, and balanced dataset $\mathcal{D}_{base}$ through data curation and automated annotation refinement.

This dataset satisfies:
\begin{equation}
\mathcal{D}_{base} \sim p_{div}(x),
\end{equation}
where $p_{div}(x)$ denotes a high-entropy data distribution covering diverse layouts, languages, document types, and visual styles.





This stage emphasizes broad visual-semantic alignment, stable feature learning, and robust cross-domain generalization. Although $\mathcal{D}_{base}$ contains moderate annotation noise, its large scale and diversity enable effective representation learning. From an optimization perspective, training on $\mathcal{D}_{base}$ yields a relatively smooth loss landscape, facilitating stable convergence.

\subsubsection{Stage II: Uncertainty-Driven Boundary Refinement}
After Stage I convergence, the model acquires strong general capabilities. However, performance is constrained by noisy supervision and limited exposure to complex edge cases. To overcome this limitation, we introduce an uncertainty-driven curriculum refinement stage.

\paragraph{Hard Case Mining via Inference Consistency.}

For each unlabeled or weakly labeled sample $x$, we perform $T$ stochastic inference passes:
\begin{equation}
\{\hat{y}^{(t)}\}_{t=1}^{T}
=
\{ f_\theta(x;\xi_t) \}_{t=1}^{T},
\end{equation}
where $\xi_t$ represents stochastic factors such as sampling temperature or dropout.

We define a task-specific consistency metric $S(\cdot,\cdot)$: (1) PageIoU for layout, (2) CDM for formula, (3) TEDS for tables.

The mean consistency score is computed as:
\begin{equation}
C(x)=
\frac{2}{T(T-1)}
\sum_{i<j}
S(\hat{y}^{(i)},\hat{y}^{(j)}).
\end{equation}

Low values of $C(x)$ indicate high prediction uncertainty.

We select hard samples as:
\begin{equation}
\mathcal{D}_{hard}=\{x \mid C(x)<\tau\},
\end{equation}
where $\tau$ is a task-dependent threshold.


\paragraph{Expert Refinement and Dataset Construction.}

Samples in $\mathcal{D}_{hard}$ are processed through an AI-assisted human annotation pipeline to produce high-precision labels:
\begin{equation}
\tilde{\mathcal{D}}_{hard}=\{x_i,\tilde{y}_i\}_{i=1}^{M}.
\end{equation}

The final fine-tuning dataset is constructed as:
\begin{equation}
\mathcal{D}_{SFT}
=
\tilde{\mathcal{D}}_{hard}
\cup
\alpha\,\mathcal{D}_{rand},
\end{equation}
where $\mathcal{D}_{rand}$ is a randomly sampled subset of $\mathcal{D}_{base}$, and $\alpha$ controls the regularization ratio.


The Stage II optimization objective is defined as:
\begin{equation}
\mathcal{L}_{hard}(\theta)
=
\mathbb{E}_{(x,y)\sim\mathcal{D}_{SFT}}
\left[
w(x)\,\ell(f_\theta(x),y)
\right],
\end{equation}

where the sample weight is:
\begin{equation}
w(x)=1+\beta(1-C(x)),
\end{equation}
with $\beta$ controlling the emphasis on hard samples.

This adaptive weighting further encourages the model to focus on decision-boundary regions. The progressive curriculum mitigates optimization instability and performance ceilings caused by diffusion’s any-order modeling by organizing data from broad to difficult, enabling \mineru to overcome annotation noise and long-tail complexity for superior real-world document OCR performance.

\section{Experiments}
\label{section:eval}
In this section, we present a comprehensive quantitative evaluation of \mineru{} to demonstrate its effectiveness in document OCR tasks. More ablation studies and qualitative examples are provided in \hyperref[app:attn]{Appendix~\ref*{app:attn}} and \hyperref[app:examples]{Appendix~\ref*{app:examples}}.

\subsection{Experimental Setups}

\subsubsection{Data}

All meta training data are derived from the MinerU2.5 dataset \cite{niu2025mineru2}, with a total volume of approximately 7.5M samples. The dataset primarily focuses on Chinese and English document parsing tasks. Therefore, no dedicated evaluation was conducted for low-resource languages.

\subsubsection{Models and Optimization}

Our experiments adopt a block-wise attention dVLM architecture. Specifically, we employ the SDAR-1.7B-Chat-b32 \cite{cheng2025sdar} with a block size of 32. The remaining components follow the MinerU2.5 architecture, except that M-RoPE \cite{wang2024qwen2, su2024roformer} is removed. We first fine-tune the \mineru{} on the LLaVA-NeXT dataset \cite{liu2023visual} for visual question answering (VQA) tasks. Based on this initialization, we further conduct specialized training for document optical character recognition (OCR). Additional optimization details are provided in \hyperref[app:training_details]{Appendix~\ref*{app:training_details}}.

\subsubsection{Evaluation}

All experiments are conducted with a block size of 32 and a dynamic decoding strategy. The decoding threshold is set to $T = 0.95$, with top-$k = 0$, temperature $= 1.0$, and top-$p = 1.0$.

For full document parsing and layout analysis, we evaluate our models on OmniDocBench v1.5 \cite{ouyang2025omnidocbench}. Table recognition is assessed using CC-OCR \cite{yang2025cc} and OCRBench v2 \cite{fu2024ocrbench}, while formula recognition is evaluated on UniMER-Test \cite{wang2024unimernet}. The inference prompts for these tasks are summarized in \hyperref[app:prompt_templates]{Appendix~\ref*{app:prompt_templates}}.

Unless otherwise stated, all OmniDocBench results use the same inference setting as above and follow the latest evaluation protocol on 1,355 pages with hybrid matching. On OmniDocBench, text is evaluated by edit distance ($\downarrow$), formulas by CDM ($\uparrow$), and tables by TEDS / TEDS-S ($\uparrow$). The Overall score is computed from the three core parsing metrics:
\[
\text{Overall} =
\frac{(1-\text{Text}^{\text{Edit}})\times 100 + \text{Formula}^{\text{CDM}} + \text{Table}^{\text{TEDS}}}{3}.
\]
Therefore, Reading Order and Table$_{\textit{TEDS-S}}$ are auxiliary metrics and are not included in Overall. Reading Order is only reported under the \textit{w/o GT Layout} setting, where the model must jointly predict layout and content from the full page. In contrast, under \textit{w/ GT Layout}, oracle layout regions are provided, so the evaluation mainly isolates recognition quality after removing layout detection errors.

\subsection{Full-Document Parsing Task Results}

\begin{table*}[t]
\centering
\small
\renewcommand{\arraystretch}{1.15}
\resizebox{\textwidth}{!}{%
\begin{tabular}{llc|cccccc|c}
\toprule
\textbf{Type} & \textbf{Methods} & \textbf{Params.} &
\textbf{Overall} $\uparrow$ &
\textbf{Text} $\downarrow$ &
\textbf{Formula} $\uparrow$ &
\textbf{Table}$_{\textit{TEDS}}$ $\uparrow$ &
\textbf{Table}$_{\textit{TEDS-S}}$ $\uparrow$ &
\textbf{Reading Order} $\downarrow$ &
\textbf{GT Layout} \\
\midrule

\multirow{2}{*}{Pipeline}
& Mineru2-pipeline~\cite{wang2024mineru}    & -     & 75.51 & 0.209 & 76.55 & 70.90 & 79.11 & 0.225 & \multicolumn{1}{c}{$\times$} \\
& PP-StructureV3~\cite{cui2025paddleocr}      & -     & 86.73 & 0.073 & 85.79 & 81.68 & 89.48 & 0.073 & \multicolumn{1}{c}{$\times$} \\
\midrule

\multirow{9}{*}{AR}
& Qwen2.5-VL-72B~\cite{qwen2.5-VL}      & 72B   & 87.02 & 0.094 & 88.27 & 82.15 & 86.22 & 0.102 & \multicolumn{1}{c}{$\times$} \\
& Gemini-2.5 Pro~\cite{comanici2025gemini}      & -     & 88.03 & 0.075 & 85.82 & 85.71 & 90.29 & 0.097 & \multicolumn{1}{c}{$\times$} \\
& MinerU2-VLM~\cite{wang2024mineru}         & 0.9B  & 85.56 & 0.078 & 80.95 & 83.54 & 87.66 & 0.086 & \multicolumn{1}{c}{$\times$} \\
& Nanonets-OCR-s~\cite{nanonets2025}      & 3B    & 85.59 & 0.093 & 85.90 & 80.14 & 85.57 & 0.108 & \multicolumn{1}{c}{$\times$} \\
& MonkeyOCR-pro-1.2B~\cite{li2025monkeyocr}  & 1.9B  & 86.96 & 0.084 & 85.02 & 84.24 & 89.02 & 0.130 & \multicolumn{1}{c}{$\times$} \\
& MonkeyOCR-pro-3B~\cite{li2025monkeyocr}    & 3.7B  & 88.85 & 0.075 & 87.25 & 86.78 & 90.63 & 0.128 & \multicolumn{1}{c}{$\times$} \\
& dots.ocr~\cite{li2025dots}            & 3B    & 88.41 & 0.048 & 83.22 & 86.78 & 90.62 & 0.053 & \multicolumn{1}{c}{$\times$} \\
& MinerU2.5~\cite{niu2025mineru2}           & 1.2B  & 90.67 & 0.047 & 88.46 & 88.22 & 92.38 & 0.044 & \multicolumn{1}{c}{$\times$} \\
& PaddleOCR-VL~\cite{cui2025paddleocr}        & 0.9B  & 92.56 & 0.035 & 91.43 & 89.76 & 93.52 & 0.043 & \multicolumn{1}{c}{$\times$} \\
\midrule

\rowcolor{green!15}
\multirow{1}{*}{dLM}
& \mineru             & 2.5B  & 88.94 & 0.061 & 86.41 & 86.50 & 90.29 & 0.059 & \multicolumn{1}{c}{$\times$} \\
\midrule
\rowcolor{white}

\multirow{2}{*}{AR}
& MinerU2.5~\cite{niu2025mineru2}           & 1.2B  & 93.44 & 0.025 & 91.98 & 90.84 & 95.10 & -     & \multicolumn{1}{c}{$\checkmark$} \\
& PaddleOCR-VL~\cite{cui2025paddleocr}        & 0.9B  & 93.91 & 0.021 & 92.13 & 91.70 & 95.42 & -     & \multicolumn{1}{c}{$\checkmark$} \\
\midrule

\rowcolor{green!15}
\multirow{1}{*}{dLM}
& \mineru             & 2.5B  & 93.37 & 0.028 & 91.92 & 91.00 & 94.86 & -     & \multicolumn{1}{c}{$\checkmark$} \\

\bottomrule
\end{tabular}%
}
\caption{Comprehensive evaluation of document parsing on OmniDocBench v1.5. $\uparrow$ denotes higher is better, $\downarrow$ denotes lower is better.}
\label{tab:full-document}
\end{table*}

As shown in \Cref{tab:full-document}, under the fully automatic setting without GT Layout, \mineru achieves an Overall score of 88.94, outperforming most AR-based models and demonstrating strong end-to-end parsing capability without relying on oracle layout information. When evaluated with GT Layout, \mineru further improves to 93.37 Overall, reaching a score that is close to top-tier AR-based systems and indicating high competitiveness in overall parsing performance. Meanwhile, the sizable gap between the two settings suggests that layout understanding remains a key bottleneck: \mineru can leverage accurate layout signals effectively, and its remaining weaknesses are primarily attributed to layout analysis, leaving clear room for improvement in layout prediction to further close the gap in fully automatic parsing.

To complement the aggregate results in \Cref{tab:full-document}, \Cref{tab:omnidocbench-by-category} further breaks down the text edit distance over the nine OmniDocBench page types. Lower values indicate better page-level parsing quality.

\FloatBarrier
\begin{table*}[!t]
\small
\centering
\setlength{\tabcolsep}{4pt}
\renewcommand{\arraystretch}{1.05}
\resizebox{\textwidth}{!}{%
\begin{tabular}{ll|ccccccccc|c}
\toprule
\textbf{Type} & \textbf{Methods} &
\textbf{Slides} &
\makecell{\textbf{Academic}\\\textbf{Papers}} &
\textbf{Book} &
\textbf{Textbook} &
\makecell{\textbf{Exam}\\\textbf{Papers}} &
\textbf{Magazine} &
\textbf{Newspaper} &
\textbf{Notes} &
\makecell{\textbf{Financial}\\\textbf{Report}} &
\makecell{\textbf{GT}\\\textbf{Layout}} \\
\midrule

\multirow{2}{*}{Pipeline}
& MinerU2-pipeline~\cite{wang2024mineru} & 0.4244 & 0.0230 & 0.2628 & 0.1224 & 0.0822 & 0.3950 & 0.0736 & 0.2603 & 0.0411 & $\times$ \\
& PP-StructureV3~\cite{cui2025paddleocr} & 0.0794 & 0.0236 & 0.0415 & 0.1107 & 0.0945 & 0.0722 & 0.0617 & 0.1236 & 0.0181 & $\times$ \\
\midrule

\multirow{9}{*}{AR}
& Qwen2.5-VL-72B~\cite{qwen2.5-VL} & 0.0422 & 0.0801 & 0.0586 & 0.1146 & 0.0681 & 0.0964 & 0.2380 & 0.1232 & 0.0264 & $\times$ \\
& Gemini-2.5 Pro~\cite{comanici2025gemini} & 0.0326 & 0.0182 & 0.0694 & 0.1618 & 0.0937 & 0.0161 & 0.1347 & 0.1169 & 0.0169 & $\times$ \\
& MinerU2-VLM~\cite{wang2024mineru} & 0.0745 & 0.0104 & 0.0357 & 0.1276 & 0.0698 & 0.0652 & 0.1831 & 0.0803 & 0.0236 & $\times$ \\
& Nanonets-OCR-s~\cite{nanonets2025} & 0.0551 & 0.0578 & 0.0606 & 0.0931 & 0.0834 & 0.0917 & 0.1965 & 0.1606 & 0.0395 & $\times$ \\
& MonkeyOCR-pro-1.2B~\cite{li2025monkeyocr} & 0.0961 & 0.0354 & 0.0530 & 0.1110 & 0.0887 & 0.0494 & 0.0995 & 0.1686 & 0.0198 & $\times$ \\
& MonkeyOCR-pro-3B~\cite{li2025monkeyocr} & 0.0879 & 0.0459 & 0.0517 & 0.1067 & 0.0726 & 0.0482 & 0.0937 & 0.1141 & 0.0211 & $\times$ \\
& dots.ocr~\cite{li2025dots} & 0.0290 & 0.0231 & 0.0433 & 0.0788 & 0.0467 & 0.0221 & 0.0667 & 0.1116 & 0.0076 & $\times$ \\
& MinerU2.5~\cite{niu2025mineru2} & 0.0304 & 0.0213 & 0.0322 & 0.0590 & 0.0501 & 0.0388 & 0.0570 & 0.1036 & 0.0071 & $\times$ \\
& PaddleOCR-VL~\cite{cui2025paddleocr} & 0.0260 & 0.0168 & 0.0320 & 0.0691 & 0.0447 & 0.0194 & 0.0357 & 0.0593 & 0.0079 & $\times$ \\
\midrule

\rowcolor{green!15}
\multirow{1}{*}{dLM}
& \mineru & 0.0440 & 0.0244 & 0.0444 & 0.0677 & 0.0580 & 0.0474 & 0.0863 & 0.1618 & 0.0104 & $\times$ \\
\midrule
\rowcolor{white}

\multirow{2}{*}{AR}
& MinerU2.5~\cite{niu2025mineru2} & 0.0164 & 0.0184 & 0.0144 & 0.0259 & 0.0273 & 0.0126 & 0.0288 & 0.0559 & 0.0045 & $\checkmark$ \\
& PaddleOCR-VL~\cite{cui2025paddleocr} & 0.0173 & 0.0224 & 0.0145 & 0.0349 & 0.0292 & 0.0129 & 0.0311 & 0.0455 & 0.0177 & $\checkmark$ \\
\midrule

\rowcolor{green!15}
\multirow{1}{*}{dLM}
& \mineru & 0.0253 & 0.0226 & 0.0306 & 0.0316 & 0.0281 & 0.0181 & 0.0336 & 0.0528 & 0.0053 & $\checkmark$ \\

\bottomrule
\end{tabular}%
}
\caption{Breakdown of document parsing on OmniDocBench v1.5 by page type, reported in text edit distance ($\downarrow$).}
\label{tab:omnidocbench-by-category}
\end{table*}

\FloatBarrier

\subsection{Element-Specific Parsing Task Results}

\subsubsection{Table Recognition}

\begin{table}[t]
\centering
\small
\setlength{\tabcolsep}{4pt}
\renewcommand{\arraystretch}{1.05}
\resizebox{\textwidth}{!}{%
\begin{tabular}{l l|cc|cc|cccc}
\toprule
\multirow{2}{*}{\textbf{Type}} & \multirow{2}{*}{\textbf{Method}}
& \multicolumn{2}{c|}{\textbf{CC-OCR}}
& \multicolumn{2}{c|}{\textbf{OCRBench v2}}
& \multicolumn{4}{c}{\textbf{UniMER-Test}} \\
& & TEDS$\uparrow$ & TEDS-S$\uparrow$
& TEDS$\uparrow$ & TEDS-S$\uparrow$
& CPE$\uparrow$ & HWE$\uparrow$ & SCE$\uparrow$ & SPE$\uparrow$ \\
\midrule
\multirow{5}{*}{AR}
& InternVL3.5-241B~\cite{wang2025internvl3}  & 62.87 & 69.52 & 79.50 & 85.81 & 91.7 & 93.2 & 95.1 & 97.8 \\
& Qwen2.5-VL-72B~\cite{qwen2.5-VL}    & 81.22 & 86.48 & 81.33 & 86.58 & 88.9 & 91.8 & 95.5 & 96.2 \\
& GPT-4o~\cite{hurst2024gpt}            & 66.98 & 79.04 & 70.51 & 79.55 & 82.7 & 85.9 & 87.8 & 96.7 \\
& dots.ocr~\cite{li2025dots}          & 75.42 & 81.65 & 82.04 & 86.27 & 86.8 & 90.5 & 94.7 & 97.5 \\
& MinerU2.5~\cite{niu2025mineru2}         & 79.76 & 85.16 & 87.13 & 90.62 & 96.6 & 94.4 & 96.4 & 98.4 \\
\midrule
\rowcolor{green!15}
dLM & \mineru         & 73.77 & 82.06 & 81.18 & 88.66 & 91.6 & 91.6 & 92.0 & 96.8 \\
\bottomrule
\end{tabular}%
}
\caption{Comprehensive recognition results on CC-OCR~\cite{yang2025cc}, OCRBench v2~\cite{fu2024ocrbench}, and UniMER-Test~\cite{wang2024unimernet}. $\uparrow$ denotes higher is better.}
\label{tab:element-specific}
\end{table}

We evaluate \mineru’s table recognition performance on the table subsets of CC-OCR~\cite{yang2025cc} and OCRBench v2~\cite{fu2024ocrbench}, as shown in \Cref{tab:element-specific}. \mineru is competitive with other AR-based models on both table subsets. It achieves 81.18/88.66 (TEDS/TEDS-S) on OCRBench v2 and 73.77/82.06 on CC-OCR, matching or surpassing several strong AR baselines in the same evaluation setting and remaining close to the top group. In particular, the strong TEDS-S scores indicate that \mineru can reliably preserve table structure during diffusive decoding, demonstrating solid end-to-end table parsing capability and overall competitiveness among AR approaches.

\subsubsection{Formula Recognition}

We evaluate formula recognition on UniMER-Test~\cite{wang2024unimernet}, with results summarized in \Cref{tab:element-specific}. \mineru achieves 91.6/91.6/92.0/96.8 on CPE/HWE/SCE/SPE, consistently demonstrating strong performance across complex, handwritten, and printed settings. Compared with other autoregressive (AR) baselines, \mineru is clearly competitive, substantially outperforming general-purpose models such as GPT-4o~\cite{hurst2024gpt}, and reaching similar ranges to strong AR systems on multiple subsets. The remaining gap to the best specialized pipeline (e.g., MinerU2.5~\cite{niu2025mineru2}) is most visible on harder categories like CPE, suggesting future improvements should focus on more precise symbol-level modeling and structure-aware decoding for complex printed expressions.

\subsection{Ablation Study}
\label{sec:ablation}

\begin{figure}[!t]
    \centering
    \includegraphics[width=\linewidth]{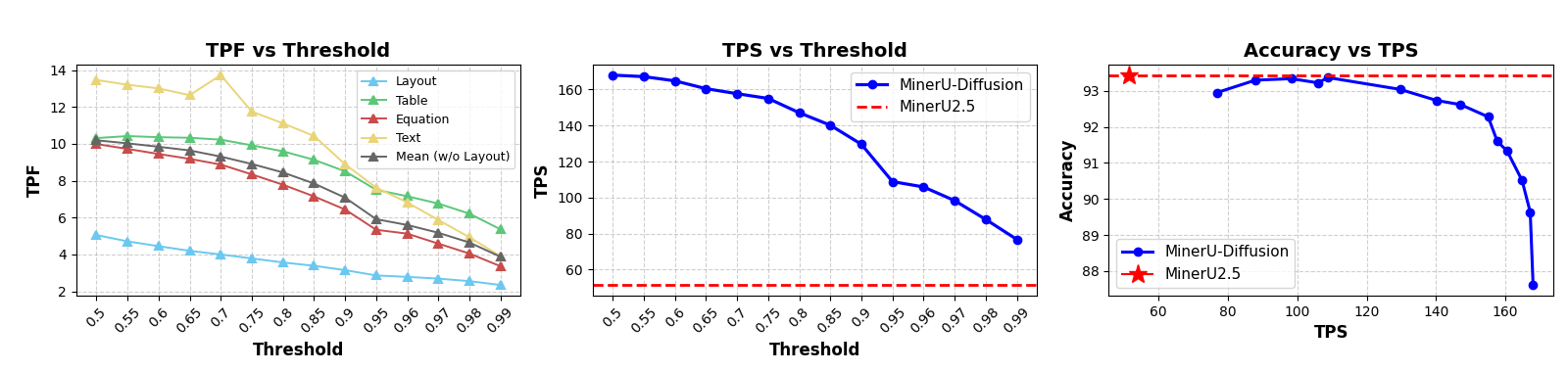}
    \caption{Threshold sensitivity analysis of TPF, TPS, and accuracy. TPF denotes tokens per forward, and TPS refers to throughput measured on an NVIDIA H200 GPU with a batch size of 1.}
    \label{fig:ab2}
\end{figure}

\subsubsection{Confidence Threshold vs. Decoding Parallelism}
\label{Confidence Threshold vs. Decoding Parallelism}

We systematically analyze the impact of dynamic confidence thresholds on decoding parallelism and inference efficiency. As shown in \Cref{fig:ab2}, as the threshold increases from 0.5 to 0.99, both Tokens Per Forward (TPF) and Throughput Per Second (TPS) decrease monotonically: lower thresholds relax confirmation constraints, enabling greater parallel decoding and higher throughput, whereas thresholds above 0.95 cause decoding to approach near token-by-token generation, leading to a marked efficiency drop. The threshold thus serves as a controllable system-level knob that enables a continuous and predictable trade-off between efficiency and conservativeness. Compared with MinerU2.5 at matched accuracy (fixed at $\sim$52 TPS), our method achieves 108.9 TPS at 93\%+ accuracy (thr = 0.95), corresponding to a 2.1$\times$ speedup, and reaches 164.8 TPS while maintaining over 90\% accuracy (thr = 0.6), yielding a peak acceleration of approximately 3.2$\times$, thereby unlocking substantial parallelism without sacrificing quality.

\subsubsection{Decoding Parallelism vs Accuracy}
\begin{figure}[!t]
    \centering
    \includegraphics[width=\linewidth]{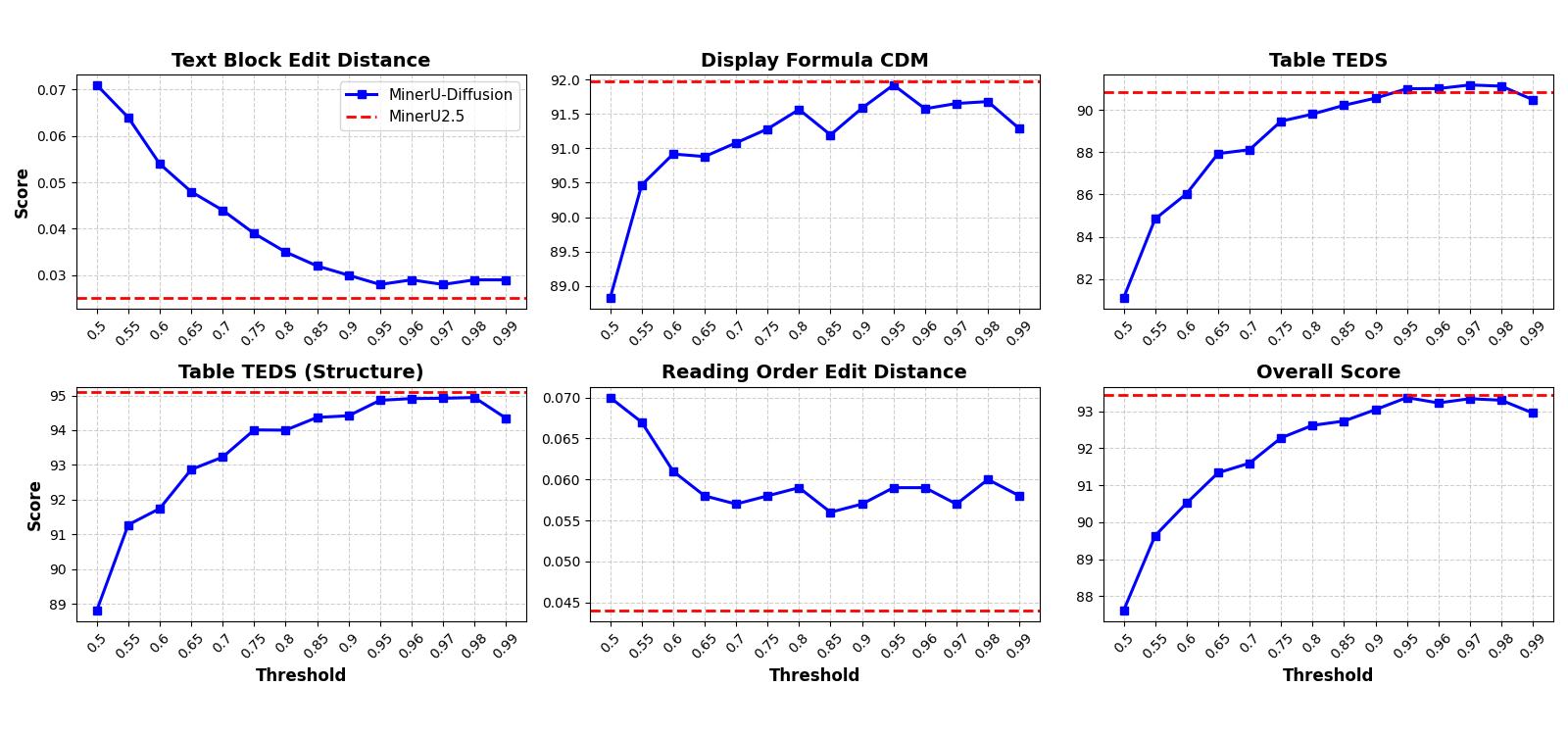}
    \caption{Visualizing the accuracy-throughput trade-off of different models across different OCR tasks under the \textit{w/ GT Layout} setting.}
    \label{fig:ab1}
\end{figure}

This experiment analyzes the impact of decoding parallelism on model accuracy by varying the dynamic confidence threshold (0.5--0.99) under the \textit{w/ GT Layout} setting. As shown in \Cref{fig:ab1}, as the threshold increases, performance first improves, then saturates, and slightly declines: the Overall score rises from 87.61 (thr=0.5) to 93.37 (thr=0.95), indicating that moderately reducing parallelism mitigates low-confidence error propagation and enhances accuracy. However, further increasing the threshold to 0.99 leads to a slight drop, suggesting that overly conservative decoding weakens the benefits of parallel information interaction. At the task level, higher thresholds consistently improve structural consistency, benefiting text, table, and formula recognition (e.g., Table TEDS improves from 81.10 to 90.99), as reduced parallelism alleviates structural error accumulation. Combined with the throughput analysis in \Cref{fig:ab2}, these results reveal a clear accuracy--efficiency trade-off: lower thresholds favor speed, while higher thresholds improve stability at the cost of efficiency. Notably, at thr=0.95, MinerU-Diffusion achieves an Overall score of 93.37, comparable to MinerU2.5 while retaining superior decoding efficiency, making it an effective balance point between accuracy and efficiency.

\subsubsection{Decoding Strategy Comparisons}

\begin{table}[t]
\centering
\small
\setlength{\tabcolsep}{3pt}
\renewcommand{\arraystretch}{1.03}
\resizebox{\textwidth}{!}{%
\begin{tabular}{l l|cc|ccccc|cc}
\toprule
\multirow{2}{*}{\textbf{Type}} 
& \multirow{2}{*}{\textbf{Method}} 
& \multicolumn{2}{c|}{\textbf{Inference}} 
& \multicolumn{5}{c|}{\textbf{Score (w/ GT Layout)}} 
& \multicolumn{2}{c}{\textbf{Score (w/o GT Layout)}} \\

& 
& TPF & TPS
& Overall$\uparrow$ 
& Text$\downarrow$
& Formula$\uparrow$ 
& Table$_{\text{TEDS}}$$\uparrow$ 
& Table$_{\text{TEDS-S}}$$\uparrow$
& Overall$\uparrow$
& Reading Order$\downarrow$ \\
\midrule

\multirow{3}{*}{dLM}
& Static (step=6)  
& 5.33 & 91.56 
& 88.31 & 0.064 & 85.52 & 85.81 & 9120
& 82.73
& 0.077 \\

& Dynamic ($\tau=0.97$) 
& 5.18 & 98.32 
& 93.34 & 0.028 & 91.65 & 91.17 & 94.92
& 89.04
& 0.057 \\

& Static (step=32) 
& 1.00 & 21.86 
& 93.02 & 0.030 & 91.10 & 90.95 & 94.73
& 89.76
& 0.060 \\

\midrule

\multirow{2}{*}{AR}
& MinerU2.5~\cite{niu2025mineru2}
& 1.00 & 51.46 
& 93.44 & 0.025 & 91.98 & 90.84 & 95.10
& 90.67
& 0.044 \\

& PaddleOCR-VL~\cite{cui2025paddleocr}
& 1.00 & 40.77 
& 93.91 & 0.021 & 92.13 & 91.70 & 95.42
& 92.79
& 0.043 \\

\bottomrule
\end{tabular}%
}
\caption{Comparison of different inference strategies and model types in terms of efficiency (TPF, TPS) and document parsing performance. }
\label{tab:inference-comparison}
\end{table}

We compare static decoding with dynamic scheduling on OmniDocBench v1.5~\cite{ouyang2025omnidocbench}, as summarized in \Cref{tab:inference-comparison}. Increasing the total decoding steps from 6 to 32 improves overall performance from 88.31 to 93.02, indicating that decoding fewer tokens per step reduces error accumulation, albeit at the cost of requiring more rounds of inference and thus lower throughput (21.86 TPS). In contrast, dynamic scheduling outperforms static step=6 in both overall and sub-task metrics (93.34 vs.\ 88.31) while achieving higher throughput (98.32 vs.\ 91.56 TPS), indicating that confidence-based adaptive token selection effectively reduces premature low-confidence commitments and balances efficiency with accuracy. As further shown in \Cref{tab:inference-comparison}, compared with strong AR baselines such as MinerU2.5~\cite{niu2025mineru2} and PaddleOCR-VL~\cite{cui2025paddleocr}, our approach achieves competitive parsing performance while maintaining high throughput, highlighting the potential of parallel generation strategies for complex document understanding.

\subsubsection{Full-Attn vs Block-Attn}
\label{sec:full-vs-block-attn}

To analyze the impact of attention mechanisms on efficiency and stability, we conducted ablation experiments on Full-Attn and Block-Attn Diffusion using LLaDA-MoE-7B-A1B-Instruct~\cite{zhu2025llada}, trained on an OCR task. The main difference lies in the amount of attention computed per forward step. Full-Attn computes roughly $L \times L$ attention, leading to quadratic growth in memory and computational costs with longer sequences, which slows inference and can induce repetitive decoding in OCR. In contrast, Block-Attn computes roughly $B \times L$ attention per forward step, using a ``self-regressive across blocks, diffusion within blocks'' design, achieving near-linear scalability while maintaining bidirectional context modeling. Experimental results show Block-Attn reduces memory and latency significantly and better mitigates repetition in OCR generation, yielding stronger scalability and stability. A more detailed discussion of these ablation results is provided in \hyperref[app:attn]{Appendix~\ref*{app:attn}}.

\subsubsection{Two-Stage Curriculum Learning}

\begin{figure}[!t]
    \centering
    \includegraphics[width=\linewidth]{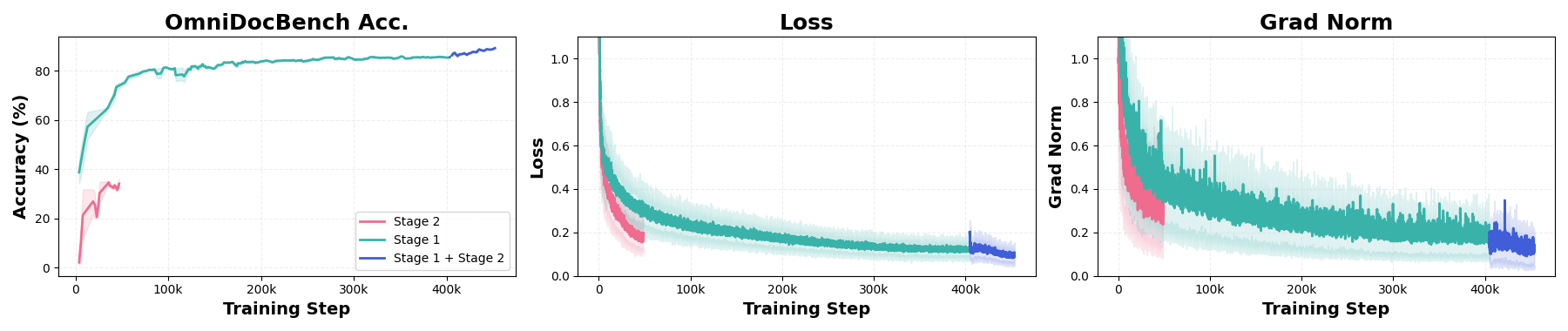}
    \caption{Comparison of training dynamics across different curriculum strategies. The two-stage framework achieves smoother optimization and higher final accuracy compared to single-stage baselines.}
    \label{fig:training}
\end{figure}

\begin{table}[t]
\centering
\resizebox{\textwidth}{!}{%
\begin{tabular}{l|ccccc|cc}
\toprule
\multirow{2}{*}{\textbf{Setting}} 
& \multicolumn{5}{c|}{\textbf{Score (w/ GT Layout)}} 
& \multicolumn{2}{c}{\textbf{Score (w/o GT Layout)}} \\

& Overall$\uparrow$
& Text$\downarrow$
& Formula$\uparrow$
& Table$_{\text{TEDS}}$$\uparrow$
& Table$_{\text{TEDS-S}}$$\uparrow$
& Overall$\uparrow$
& Reading Order$\downarrow$ \\
\midrule

Stage 1 
& 92.89 & 0.032 & 91.58 & 90.28 & 94.93 
& 86.13 & 0.056 \\

Stage 2 
& 89.33 & 0.038 & 88.50 & 83.29 & 88.93 
& 35.71 & 0.472 \\

Stage 1 + Stage 2 
& 93.37 & 0.028 & 91.92 & 91.00 & 94.86 
& 88.94 & 0.059 \\

\bottomrule
\end{tabular}%
}
\caption{Comparison of different training stages under GT and non-GT layout settings.}
\label{tab:stage-comparison}
\end{table}

To validate the proposed two-stage curriculum learning framework, we conduct systematic ablations on OmniDocBench v1.5, comparing Stage 1 only, Stage 2 only, and the full two-stage strategy. As shown in \Cref{fig:training}, training dynamics (Accuracy, Loss, and Grad Norm) reveal that Stage 2 alone exhibits rapid early loss reduction but low accuracy and significant gradient oscillation, indicating high-variance optimization caused by emphasizing high-uncertainty samples without stable representation initialization. In contrast, Stage 1 achieves smooth and stable convergence with accuracy improving to around 85\%, yet performance saturates due to limited boundary refinement. The full two-stage strategy integrates both advantages: Stage 1 establishes low-variance global representations under diverse data, while Stage 2 performs uncertainty-driven boundary refinement, enabling continued improvement beyond the Stage 1 plateau and achieving over 89\% accuracy. As shown in \Cref{tab:stage-comparison}, under the challenging \textit{w/o gt\_layout} setting, Stage 2 alone largely fails (35.712), whereas the full framework reaches 88.937, confirming that the progressive curriculum effectively mitigates optimization instability and performance ceilings induced by diffusion’s any-order modeling.

\subsection{Semantic Shuffle Analysis}
\label{subsec:semantic-shuffle}

\begin{figure}[!t]
    \centering
    \includegraphics[width=\linewidth]{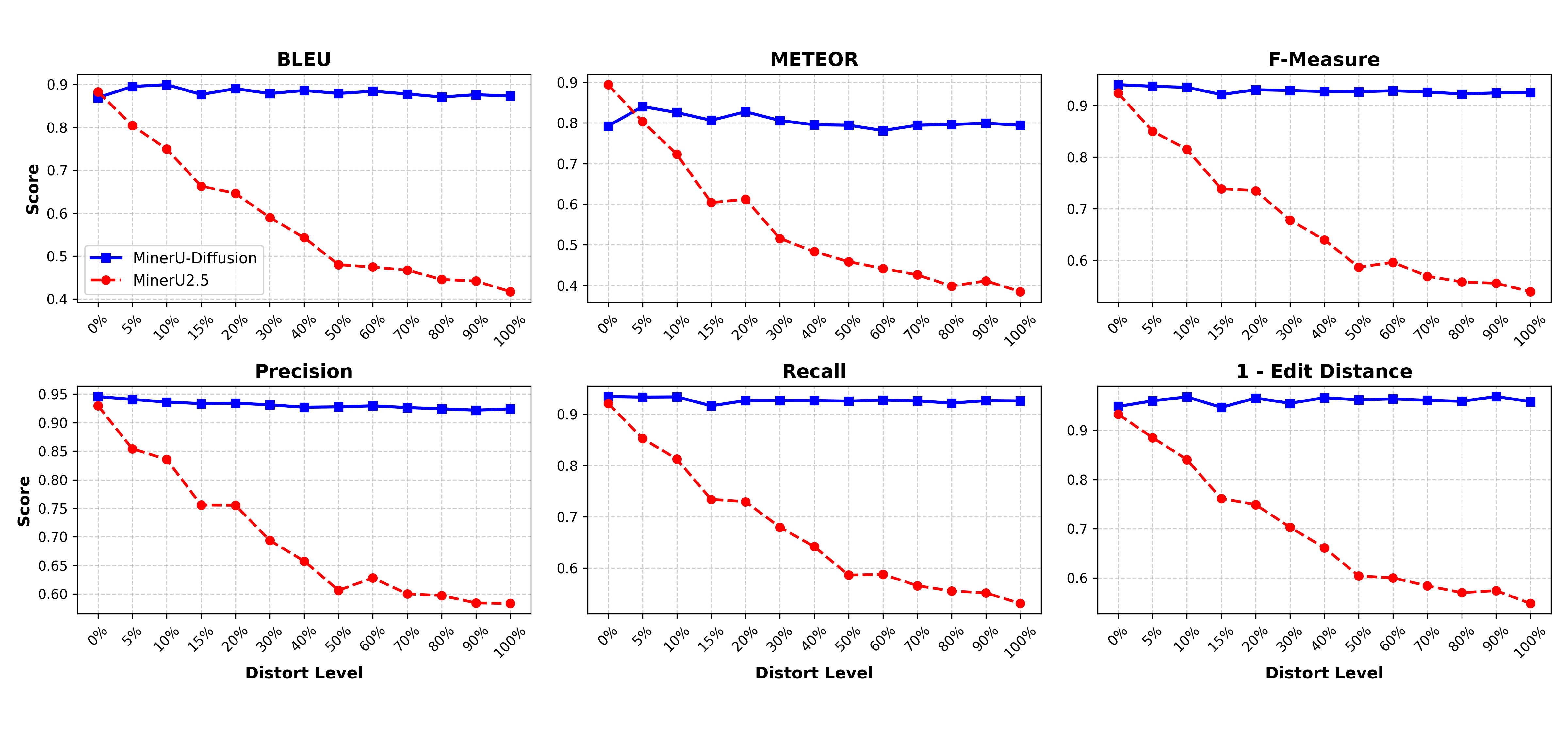}
    \caption{Semantic Shuffle benchmark results across distortion levels.}
    \label{fig:semantic-shuffle}
\end{figure}

Most modern VLM-based OCR systems \cite{niu2025mineru2, cui2025paddleocr, li2025dots, wei2025deepseek} follow a ``vision encoder + autoregressive (AR) language decoder'' pipeline: the image is converted into visual tokens, and an LM-style decoder \cite{team2024qwen2} generates the transcription token by token. Although this paradigm achieves strong OCR scores, it is often unclear whether these gains come from faithful visual reading or from the decoder's ability to ``fill in'' plausible text using linguistic priors and global context.

To better isolate true visual decoding ability, and to validate our hypothesis, we construct \textbf{Semantic Shuffle}, a benchmark that removes semantic coherence while keeping the visual presentation comparable. Starting from 112 English document images in the FOX dataset~\cite{liu2024focus}, we shuffle a controlled proportion of words and re-render the resulting text into new document images with consistent formatting.

As shown in \Cref{fig:semantic-shuffle}, experiments reveal a clear contrast between decoding paradigms. As the distortion level increases, the performance of AR decoders degrades sharply and consistently, indicating a heavy reliance on linguistic plausibility. In contrast, a diffusion-style decoder remains robust, maintaining nearly constant performance across distortion levels. These results suggest that AR decoders tend to reconstruct text using global language regularities when semantics are disrupted, whereas diffusion-based decoding aligns more directly with the visual signal under semantically invalid inputs.

\section{Conclusion}

In this paper, we propose MinerU-Diffusion, a 2.5B-parameter diffusion-based framework for document OCR, replacing autoregressive decoding with block-level parallel diffusion decoding and confidence-guided scheduling to improve efficiency and scalability. A two-stage curriculum learning strategy stabilizes training and enhances boundary precision and robustness. Experiments across document, table, formula, and Semantic Shuffle benchmarks show competitive performance against AR baselines and stronger resilience to disrupted semantics, highlighting diffusion-based parallel decoding as a promising alternative for document ocr. We hope our work inspires future research on efficient and reliable diffusion-based decoding for document OCR.

\clearpage
\newpage
\bibliographystyle{plainnat}
\setcitestyle{numbers}
\bibliography{paper}

\clearpage
\newpage
\beginappendix
\section{Training Details}
\label{app:training_details}

\mineru uses a vision encoder initialized from Qwen2-VL-7B, a diffusion decoder initialized from SDAR-1.7B-Chat-b32, and an abstractor initialized from random parameters, and is trained in three phases summarized in \Cref{tab:training_strategy}: Stage-0 initializes multimodal alignment on generic image-text data, Stage-1 performs large-scale OCR adaptation on the easier subset $\mathcal{D}_{\text{base}}$, and Stage-2 further fine-tunes the model on the harder subset $\mathcal{D}_{\text{hard}}$.

\begin{table}[htp]
    \centering
    \setlength{\tabcolsep}{12pt}
    \renewcommand{\arraystretch}{1.2}
    \resizebox{\textwidth}{!}
    { 
    \begin{tabular}{@{}ll|c|c|c|c@{}}
    \toprule
    & & \multicolumn{2}{c|}{\textbf{Stage-0}} & \textbf{Stage-1} & \textbf{Stage-2} \\ \cmidrule(l){3-4}
    & & \textbf{a} & \textbf{b} & & \\
    \midrule 
    \multirow{2}{*}{\rotatebox[origin=c]{90}{\footnotesize \textit{Vision}}}
    & \textbf{Max Resolution}   
        & $2048 \times 28 \times 28$
        & $4096 \times 28 \times 28$
        & $2048 \times 28 \times 28$
        & $2048 \times 28 \times 28$ \\
    & \#Tokens per Image 
        & $4 \sim 2048$ 
        & $4 \sim 4096$  
        & $4 \sim 2048$ 
        & $4 \sim 2048$  \\
    \midrule 
    \multirow{2}{*}{\rotatebox[origin=c]{90}{\footnotesize \textit{Data}}}
    & \textbf{Dataset} 
        & LLaVA-Pretrain
        & LLaVA-NeXT-Data
        & Layout\&OCR
        & Layout\&OCR \\
    & \#Samples 
        & 550K 
        & 739K 
        & 6.9M 
        & 630K \\
    \midrule
    \multirow{3}{*}{\rotatebox[origin=c]{90}{\footnotesize \textit{Model}}}
    & \textbf{Trainable} 
        & MLP Adaptor 
        & All 
        & All 
        & All \\
    & \textbf{Sequence Length}
        & 4096 
        & 8192 
        & 12288 
        & 16384 \\
    & \textbf{Data Augmentation}
        & No 
        & No 
        & Yes 
        & Yes \\
    \midrule 
    \multirow{3}{*}{\rotatebox[origin=c]{90}{\footnotesize \textit{Training}}}
    & \textbf{Batch Size} 
        & 256 
        & 64 
        & 256 
        & 256 \\
    & \textbf{LR} 
        & 1 $\times 10^{-3}$ 
        & 4 $\times 10^{-5}$ 
        & 4 $\times 10^{-5}$ 
        & 2 $\times 10^{-5}$ \\    
    & \textbf{Warmup Ratio}
        & 0.03 
        & 0.03
        & 0.03
        & 0.1 \\
    & \textbf{Epoch} 
        & 1 & 1 & 9 & 4 \\
    \bottomrule
    \end{tabular}
    } 
    \caption{Training setup and hyperparameters in three training stages.} 
    \label{tab:training_strategy}
\end{table}

\subsection{Stage-0: Modality Alignment}
Stage-0 provides the initialization used before document-specific supervision is introduced. In Stage-0a, LLaVA-Pretrain is used to optimize only the MLP adaptor, while the remaining modules are frozen. This step aligns visual features with the language model input space and stabilizes subsequent OCR training. In Stage-0b, training is continued on LLaVA-NeXT-Data with all parameters unfrozen, improving multimodal instruction following and adaptation to longer visual contexts.

As shown in \Cref{tab:training_strategy}, this phase uses shorter sequences and no document augmentation. Its primary role is to provide a stable initialization for later large-scale OCR optimization.

\subsection{Stage-1: Large-Scale OCR Adaptation}
Stage-1 constitutes the main OCR training phase and corresponds to the easier split $\mathcal{D}_{\text{base}}$ in \Cref{sec:two-stage-curriculum}. Training is conducted on the Layout\&OCR mixture, which jointly covers page-level layout detection and element-level recognition for text, formulas, and tables. Layout samples use full-page images with relative coordinates, whereas recognition samples use cropped regions to preserve local detail without incurring the full-page token cost.

The objective of this phase is to establish reliable OCR performance on standard document samples before emphasizing more difficult cases. Because masked diffusion is trained from partially observed targets rather than a fixed left-to-right prefix, optimization is more sensitive to label noise and heterogeneous data quality. Stage-1 therefore relies on scale, task diversity, and document augmentation to improve training stability. All parameters are updated in this phase. The sequence length is 12288, the batch size is 256, the learning rate is $4 \times 10^{-5}$, and training lasts for 9 epochs.

\subsection{Stage-2: Hard-Case Specialization}
Stage-2 starts from the Stage-1 checkpoint and shifts the sampling distribution toward the harder subset $\mathcal{D}_{\text{hard}}$. This subset contains samples on which the model remains comparatively unstable, including crowded layouts, ambiguous crop boundaries, structurally complex tables, and noisy supervision. Hard samples are mixed with a smaller replay portion drawn from the broader Stage-1 distribution so that baseline OCR coverage is preserved while failure cases are further reduced.

This phase is intended to improve performance on the samples that dominate the residual error after Stage-1. Accordingly, the sequence length is increased to 16384 and the learning rate is reduced to $2 \times 10^{-5}$, while the batch size is kept at 256 and the warmup ratio is increased to 0.1. All parameters remain trainable, and fine-tuning lasts for 4 epochs.

\subsection{Training Dynamics}

\begin{figure}[!t]
    \centering
    \includegraphics[width=\linewidth]{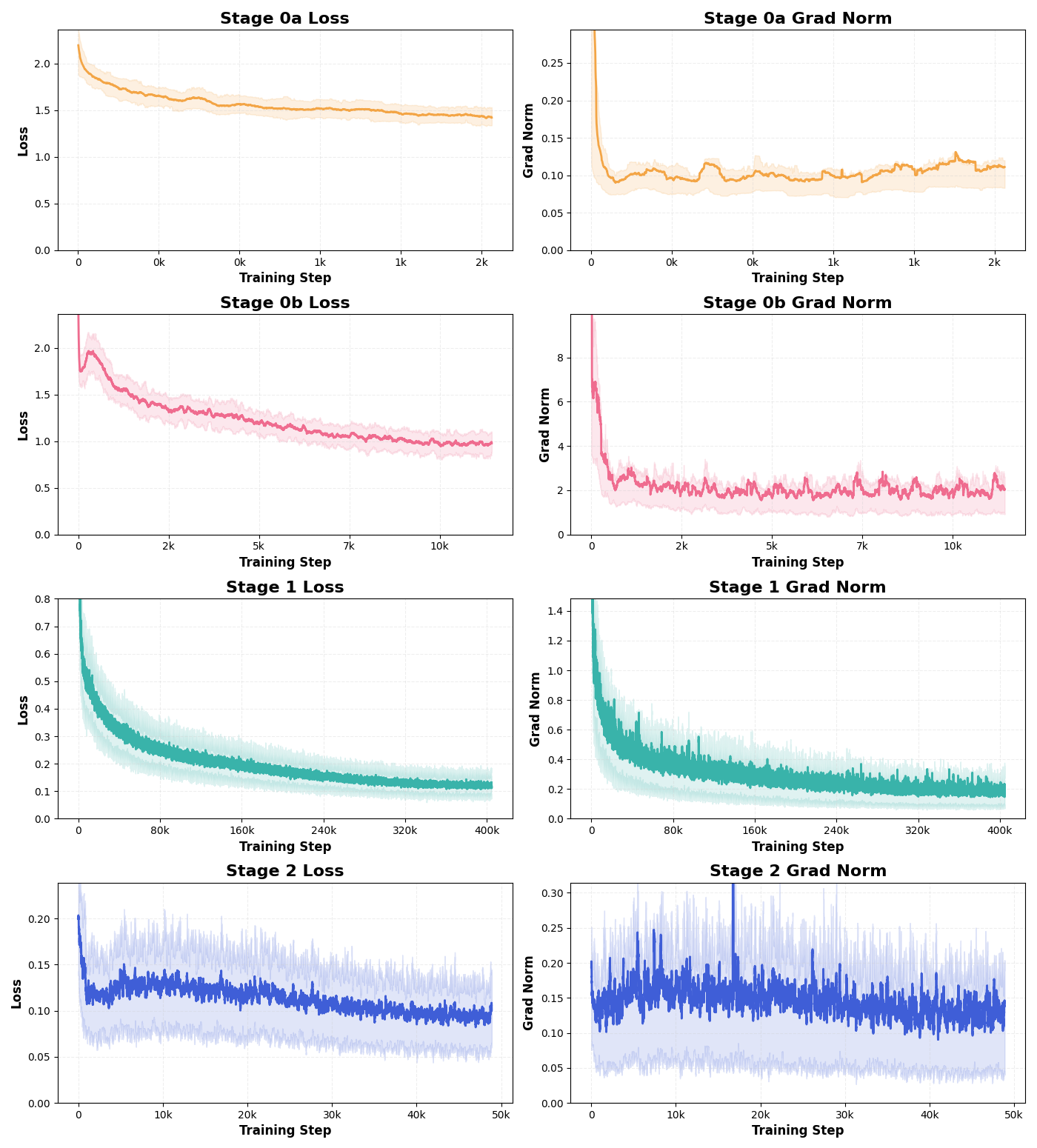}
    \caption{Training dynamics across the four stages in our training recipe: Stage~0a, Stage~0b, Stage~1, and Stage~2. For each stage, we report both the loss and the gradient norm.}
    \label{fig:training_dynamics}
\end{figure}

As shown in \Cref{fig:training_dynamics}, the loss and gradient norm remain stable throughout Stage~0a, Stage~0b, Stage~1, and Stage~2. Stage~0a and Stage~0b converge rapidly during multimodal initialization, Stage~1 exhibits smooth long-horizon optimization on $\mathcal{D}_{\text{base}}$, and Stage~2 remains stable after introducing hard-case refinement. These dynamics are consistent with the curriculum design: a stable base solution is first established on broad OCR data, after which hard-case refinement improves performance while preserving optimization stability.

\subsection{Data Augmentation}
Document-specific augmentation is enabled in both Stage-1 and Stage-2. Following the same augmentation taxonomy used in MinerU2.5, we inject four types of perturbations during OCR training: geometric changes, background disturbances, color shifts, and image degradation. In practice, this includes scaling, grid distortion, and rotation; background effects such as texture overlays, weather-like artifacts, watermarking, scanlines, and shadows; photometric changes such as brightness/contrast adjustment, illumination variation, and RGB shift; and degradation operators including PSF blur, motion-like blur, Gaussian blur, and light morphological corruption such as erosion or dilation.

These transformations target common OCR noise. We mainly simulate uneven lighting, compression artifacts, scan traces, and cluttered backgrounds. As in MinerU2.5, spatial augmentation is not applied to layout detection samples, since it can alter the geometric supervision itself. Augmentation strength and sampling probability are instead tuned by task type, so that the model gains robustness without drifting too far from valid OCR targets.

\subsection{Practical Notes}
For Stage-1 and Stage-2, the visual token budget is limited to 2048 tokens per image, and inputs are resized only when this budget is exceeded. Page-level layout detection and crop-level recognition are trained separately within the same unified model. This design aligns training with downstream usage: global structure is learned from full pages, whereas precise transcription and structure recovery are learned from native-content regions.

\section{Prompt Templates}
\label{app:prompt_templates}

This section summarizes the prompts used by \mineru{} at inference time. All OCR tasks share the same system prompt:

\begin{lstlisting}[style=mystyle,numbers=none]
System: You are a helpful assistant.
\end{lstlisting}

\subsection{Layout Detection}
Layout detection uses the following prompt call:

\begin{lstlisting}[style=mystyle,numbers=none]
System: You are a helpful assistant.
User: Layout Detection:
\end{lstlisting}

The input page image is resized to $1036 \times 1036$ before decoding. The returned sequence lists document elements in reading order. Each entry serializes the relative box coordinates, the semantic category, and the orientation tag, so that the page layout can be reconstructed directly from the decoded string. A representative prediction is shown below:

\begin{lstlisting}[style=mystyle,numbers=none]
<|box_start|>130 071 368 102<|box_end|><|ref_start|>title<|ref_end|><|rotate_up|>
<|box_start|>275 068 320 072<|box_end|><|ref_start|>text<|ref_end|><|rotate_up|>
<|box_start|>015 300 496 524<|box_end|><|ref_start|>image<|ref_end|><|rotate_up|>
\end{lstlisting}

\subsection{Text Recognition}
Text recognition uses the following prompt call:

\begin{lstlisting}[style=mystyle,numbers=none]
System: You are a helpful assistant.
User: Text Recognition:
\end{lstlisting}

The image is processed at its original resolution whenever possible, with the visual token budget constrained to 4--2048 tokens. If this limit is exceeded, the image is resized accordingly. The model returns a plain text transcription without additional structural markers. A representative prediction is shown below:

\begin{lstlisting}[style=mystyle,numbers=none]
We believe that LLMs can be trained to identify proof issues without reference solutions.
\end{lstlisting}

\subsection{Formula Recognition}
Formula recognition uses the following prompt call:

\begin{lstlisting}[style=mystyle,numbers=none]
System: You are a helpful assistant.
User: Formula Recognition:
\end{lstlisting}

The image remains at native resolution unless the visual token count exceeds the range of 4--2048 tokens, in which case resizing is applied. The decoded result is a LaTeX string representing the target expression. A representative prediction is shown below:

\begin{lstlisting}[style=mystyle,numbers=none]
\[
\begin{aligned}
\frac{d\mathbf{u}}{dt} &= \mathbf{P}^{-1}\mathbf{A}\mathbf{P}\mathbf{u} \\
\alpha &= \mathrm{tr}(\mathbf{A}) = \lambda_{1} + \lambda_{2} \\
\beta &= |\mathbf{A}| = \lambda_{1}\lambda_{2}
\end{aligned}
\]
\end{lstlisting}

\subsection{Table Recognition}
Table recognition uses the following prompt call:

\begin{lstlisting}[style=mystyle,numbers=none]
System: You are a helpful assistant.
User: Table Recognition:
\end{lstlisting}

The original image resolution is preserved when the visual token count remains within 4--2048; otherwise, the image is proportionally rescaled. The model outputs the table in OTSL (Open Table Structure Language), which can be parsed into a structured table representation. A representative prediction is shown below:

\begin{lstlisting}[style=mystyle,numbers=none]
<fcel>Team<fcel>Main Score<fcel>SROCC<fcel>PLCC<nl>
<fcel>TB-VQA(ours)<fcel>0.8576<fcel>0.8493<fcel>0.8659<nl>
<fcel>2nd<fcel>0.8396<fcel>0.8408<fcel>0.8383<nl>
<fcel>3rd<fcel>0.8289<fcel>0.8261<fcel>0.8317<nl>
<fcel>4th<fcel>0.8199<fcel>0.8163<fcel>0.8236<nl>
<fcel>5th<fcel>0.7994<fcel>0.7962<fcel>0.8026<nl>
\end{lstlisting}

\section{Full-Attn vs Block-Attn}
\label{app:attn}

This section expands \Cref{sec:full-vs-block-attn} and explains why Block-Attn is consistently preferable to Full-Attn for document OCR diffusion decoding.

\subsection{Why Full-Attn Is Inferior}

We summarize the main disadvantages of Full-Attn in three aspects:
\begin{itemize}
    \item \textbf{Lower speed in both training and inference.} With output length $L$ and block size $B$, Full-Attn computes global pairwise attention over the entire target sequence, leading to quadratic attention cost per denoising step, whereas Block-Attn reduces this cost to a near-linear form in practice.
    During training, the effective context updated at each step also differs substantially. Following our implementation, Full-Attn processes roughly $L^2/B$ token interactions per step, whereas Block-Attn only needs about $2L$, which makes Block-Attn noticeably faster to train under the same block size.
    \item \textbf{Fixed-length mismatch wastes computation.} Full-Attn must predefine a global decoding length $L$, which is hard to match across documents with very different output lengths. When $L$ is much larger than the actual target length, many decoding positions are allocated but never needed. When $L$ is too small, the preset budget is exhausted before the target is completed. In both cases, the mismatch leads to inefficient use of computation.
    \item \textbf{Fixed-length mismatch also harms generation quality.} Empirically, Full-Attn works best when the preset length $L$ is close to the actual target length $N$. When $L < N$, the model may stop early or become unstable near the tail. When $L \gtrsim N$, generation is relatively stable. When $L \gg N$, the model often fails to stop cleanly and keeps producing repeated rows or empty cells.
\end{itemize}

\subsection{Generation Quality under Length Mismatch}

\begin{figure}[H]
    \centering
    \includegraphics[width=\linewidth]{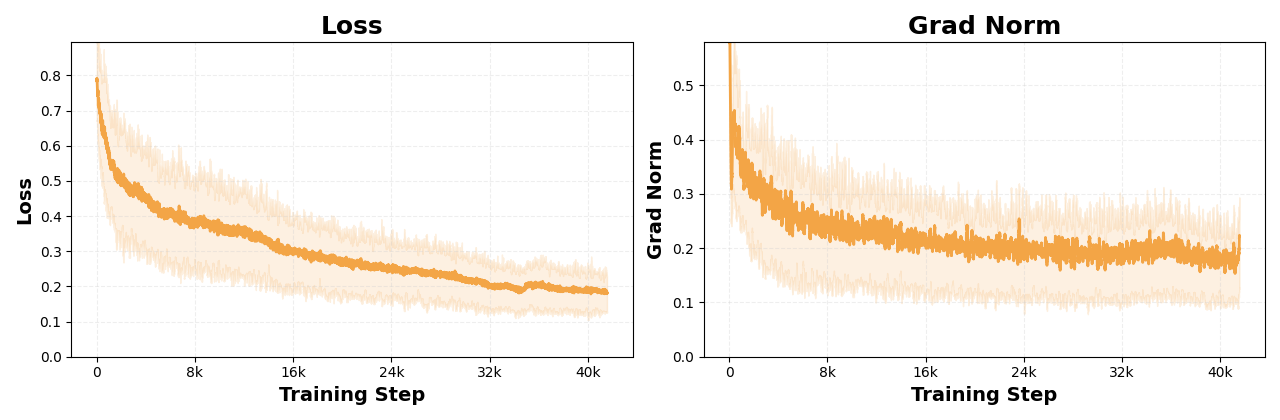}
    \caption{Training curves on a table-only subset.}
    \label{fig:full_attn_loss_grad}
\end{figure}

We next focus on the generation-quality issue caused by fixed-length mismatch in Full-Attn. As a reference, \Cref{fig:full_attn_loss_grad} shows the training curves of Full-Attn on a table-only subset; its training loss exhibits a trend consistent with Block-Attn.

We then provide three qualitative cases beyond the summary discussion in \Cref{sec:full-vs-block-attn}. We select three tables with different lengths, namely a short table, a medium-length table, and a long table. For each case, we compare Full-Attn with preset lengths $L=256$, $512$, and $1024$, together with Block-Attn at $L=1024$.

For readability, green text denotes normal content and red text denotes wrong content. The notation ``...'' denotes omitted intermediate content, ``...\texttt{\char60 repeat for $x$ times\char62}'' denotes repeated content, and \textcolor{red}{``\texttt{\char60 missing\char62}''} marks missing tail content.

\begin{figure}[H]
    \centering
    \includegraphics[width=0.50\textwidth]{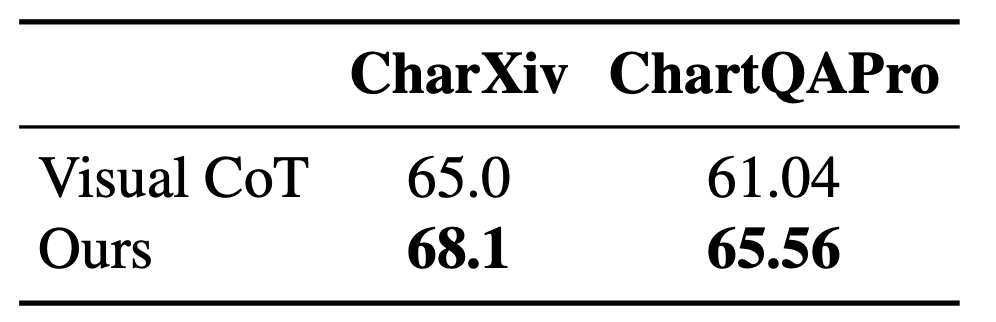}

    \begin{minipage}[t]{0.48\linewidth}
    \textbf{Full-Attn ($L=256$)}
    \begin{lstlisting}[style=attncase]
[[g]]<ecel><fcel>CharXiv<fcel>ChartQAPro<nl>[[/g]]
[[g]]<fcel>Visual CoT<fcel>65.0<fcel>61.04<nl>[[/g]]
[[g]]<fcel>Ours<fcel>68.1<fcel>65.56<nl>[[/g]]
[[r]]<ecel><ecel><ecel><nl>[[/r]]
[[r]]...<repeat for 13 times>[[/r]]
[[r]]<ecel><ecel><ecel><nl>[[/r]]
    \end{lstlisting}
    \end{minipage}\hfill
    \begin{minipage}[t]{0.48\linewidth}
    \textbf{Full-Attn ($L=512$)}
    \begin{lstlisting}[style=attncase]
[[g]]<ecel><fcel>CharXiv<fcel>ChartQAPro<nl>[[/g]]
[[g]]<fcel>Visual CoT<fcel>65.0<fcel>61.04<nl>[[/g]]
[[g]]<fcel>Ours<fcel>68.1<fcel>65.56<nl>[[/g]]
[[r]]<ecel><ecel><ecel><nl>[[/r]]
[[r]]...<repeat for 33 times>[[/r]]
[[r]]<ecel><ecel><ecel><nl>[[/r]]
    \end{lstlisting}
    \end{minipage}


    \begin{minipage}[t]{0.48\linewidth}
    \textbf{Full-Attn ($L=1024$)}
    \begin{lstlisting}[style=attncase]
[[g]]<ecel><fcel>CharXiv<fcel>ChartQAPro<nl>[[/g]]
[[g]]<fcel>Visual CoT<fcel>65.0<fcel>61.04<nl>[[/g]]
[[g]]<fcel>Ours<fcel>68.1<fcel>65.56<nl>[[/g]]
[[r]]<ecel><ecel><ecel><nl>[[/r]]
[[r]]...<repeat for 72 times>[[/r]]
[[r]]<ecel><ecel><ecel><nl>[[/r]]
    \end{lstlisting}
    \end{minipage}\hfill
    \begin{minipage}[t]{0.48\linewidth}
    \textbf{Block-Attn ($L=1024$)}
    \begin{lstlisting}[style=attncase]
[[g]]<ecel><fcel>CharXiv<fcel>ChartQAPro<nl>[[/g]]
[[g]]<fcel>Visual CoT<fcel>65.0<fcel>61.04<nl>[[/g]]
[[g]]<fcel>Ours<fcel>68.1<fcel>65.56<nl>[[/g]]
    \end{lstlisting}
    \end{minipage}

    \caption{A very short table where over-provisioned Full-Attn repeats empty rows.}
    \label{fig:full_attn_case1}
\end{figure}

\begin{figure}[H]
    \centering
    \includegraphics[width=0.58\textwidth]{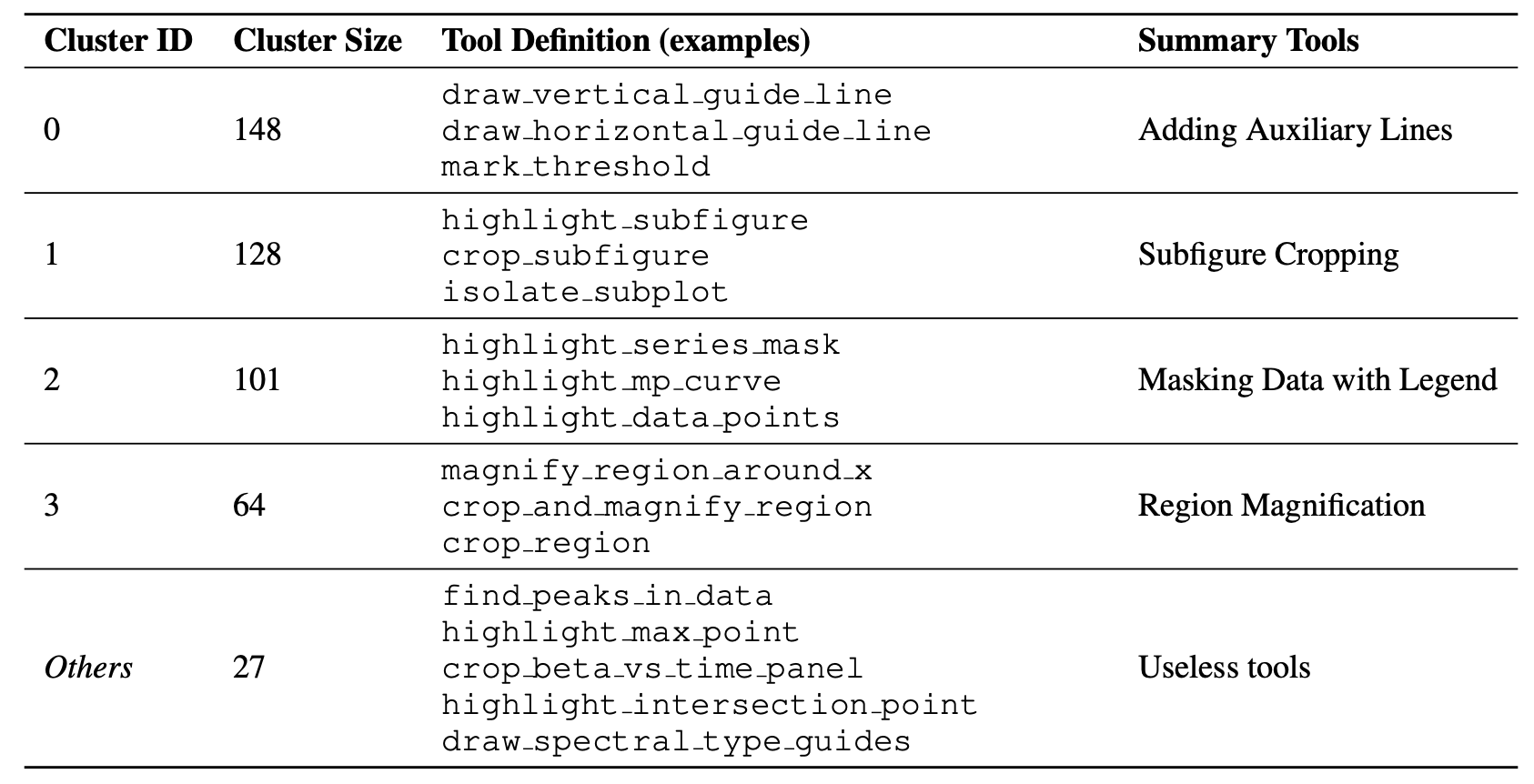}

    \begin{minipage}[t]{0.48\linewidth}
    \textbf{Full-Attn ($L=256$)}
    \begin{lstlisting}[style=attncase]
[[g]]<fcel>Cluster ID<fcel>Cluster Size...<nl>[[/g]]
[[g]]<fcel>0<fcel>148...Adding Auxiliary Lines<nl>[[/g]]
...
[[g]]<fcel>3<fcel>64...Region Magnification<nl>[[/g]]
[[g]]<fcel>Others<fcel>27...Useless tools<nl>[[/g]]
    \end{lstlisting}
    \end{minipage}\hfill
    \begin{minipage}[t]{0.48\linewidth}
    \textbf{Full-Attn ($L=512$)}
    \begin{lstlisting}[style=attncase]
[[g]]<fcel>Cluster ID<fcel>Cluster Size...<nl>[[/g]]
[[g]]<fcel>0<fcel>148...Adding Auxiliary Lines<nl>[[/g]]
...
[[g]]<fcel>3<fcel>64...Region Magnification<nl>[[/g]]
[[g]]<fcel>Others<fcel>27...Useless tools<nl>[[/g]]
    \end{lstlisting}
    \end{minipage}


    \begin{minipage}[t]{0.48\linewidth}
    \textbf{Full-Attn ($L=1024$)}
    \begin{lstlisting}[style=attncase]
[[g]]<fcel>Cluster ID<fcel>Cluster Size...<nl>[[/g]]
[[g]]<fcel>0<fcel>148<fcel>draw_vertical_guide_line...<nl>[[/g]]
...
[[g]]<fcel>Others<fcel>27...Useless tools<nl>[[/g]]
[[r]]<ucel><ecel><ucel><nl>[[/r]]
[[r]]...<repeat for 47 times>[[/r]]
[[r]]<ucel><ecel><ucel><nl>[[/r]]
    \end{lstlisting}
    \end{minipage}\hfill
    \begin{minipage}[t]{0.48\linewidth}
    \textbf{Block-Attn ($L=1024$)}
    \begin{lstlisting}[style=attncase]
[[g]]<fcel>Cluster ID<fcel>Cluster Size...<nl>[[/g]]
[[g]]<fcel>0<fcel>148...Adding Auxiliary Lines<nl>[[/g]]
...
[[g]]<fcel>Others<fcel>27...Useless tools<nl>[[/g]]
    \end{lstlisting}
    \end{minipage}

    \caption{A medium-length table where matched Full-Attn remains stable but over-provisioned Full-Attn repeats.}
    \label{fig:full_attn_case2}
\end{figure}

\begin{figure}[H]
    \centering
    \includegraphics[width=0.58\textwidth]{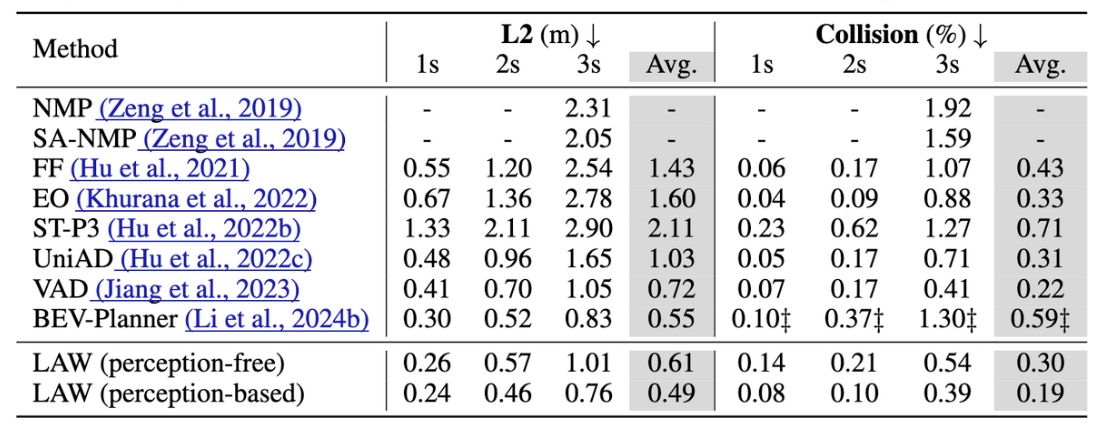}

    \begin{minipage}[t]{0.48\linewidth}
    \textbf{Full-Attn ($L=256$)}
    \begin{lstlisting}[style=attncase]
[[g]]<fcel>Method<fcel>L2 (m) ...<nl>[[/g]]
[[g]]<fcel>NMP (Zeng et al., 2019)...<nl>[[/g]]
[[g]]<fcel>SA-NMP (Zeng et al., 2019)...<nl>[[/g]]
[[g]]<fcel>FF (Hu et al., 2021)<fcel>0.55...<nl>[[/g]]
[[r]]<missing>[[/r]]
    \end{lstlisting}
    \end{minipage}\hfill
    \begin{minipage}[t]{0.48\linewidth}
    \textbf{Full-Attn ($L=512$)}
    \begin{lstlisting}[style=attncase]
[[g]]<fcel>Method<fcel>L2 (m) ...<nl>[[/g]]
...
[[g]]<fcel>ST-P3 (Hu et al., 2022b)...<nl>[[/g]]
[[g]]<fcel>UniAD (Hu et al., 2022c)...<nl>[[/g]]
[[r]]<missing>[[/r]]
    \end{lstlisting}
    \end{minipage}


    \begin{minipage}[t]{0.48\linewidth}
    \textbf{Full-Attn ($L=1024$)}
    \begin{lstlisting}[style=attncase]
[[g]]<fcel>Method<fcel>L2 (m) ...<nl>[[/g]]
...
[[g]]<fcel>LAW (perception-free)...<nl>[[/g]]
[[g]]<fcel>LAW (perception-based)...<nl>[[/g]]
[[r]]<fcel>LAW (perception-based)...<nl>[[/r]]
[[r]]<fcel>LAW total...<nl>[[/r]]
    \end{lstlisting}
    \end{minipage}\hfill
    \begin{minipage}[t]{0.48\linewidth}
    \textbf{Block-Attn ($L=1024$)}
    \begin{lstlisting}[style=attncase]
[[g]]<fcel>Method<fcel>1s<fcel>L2 (m) ...<nl>[[/g]]
...
[[g]]<fcel>VAD (Jiang et al., 2023)...<nl>[[/g]]
[[g]]<fcel>BEV-Planner Li et al., 2024b)...<nl>[[/g]]
[[g]]<fcel>LAW (perception-free)...<nl>[[/g]]
[[g]]<fcel>LAW (perception-based)...<nl>[[/g]]
    \end{lstlisting}
    \end{minipage}

    \caption{A long table where short Full-Attn truncates and long Full-Attn repeats tail rows.}
    \label{fig:full_attn_case3}
\end{figure}

For the short table in \Cref{fig:full_attn_case1}, all tested Full-Attn lengths are already over-provisioned: $L=256$ and $L=512$ produce repeated empty rows, while $L=1024$ repeats much more severely. For the medium-length table in \Cref{fig:full_attn_case2}, both $L=256$ and $L=512$ generate normally, whereas $L=1024$ introduces repeated rows after the valid content. For the long table in \Cref{fig:full_attn_case3}, $L=256$ and $L=512$ are both truncated, while $L=1024$ reaches the tail but still repeats the final record.

\lstdefinestyle{attncase}{
    style=mystyle,
    basicstyle=\ttfamily\tiny,
    numbers=none,
    literate={<}{{{\ttfamily\char60}}}1 {>}{{{\ttfamily\char62}}}1,
    moredelim=**[is][\color{ForestGreen}]{[[g]]}{[[/g]]},
    moredelim=**[is][\color{BrickRed}]{[[r]]}{[[/r]]}
}

\section{Qualitative Examples}
\label{app:examples}

\setlength{\fboxsep}{0pt}
\setlength{\fboxrule}{0.3pt}
\newcommand{\qualimg}[2]{%
\fbox{\includegraphics[width=\dimexpr\linewidth-2\fboxsep-2\fboxrule\relax,height=#2,keepaspectratio]{#1}}}
\newcommand{\qualimgfit}[2]{%
\fbox{\includegraphics[width=\dimexpr\linewidth-2\fboxsep-2\fboxrule\relax,height=#2,keepaspectratio]{#1}}}
\newcommand{\qualimgwidth}[1]{%
\fbox{\includegraphics[width=\dimexpr\linewidth-2\fboxsep-2\fboxrule\relax]{#1}}}
\newcommand{\qualpanel}[1]{%
\begin{minipage}[c][0.27\textheight][c]{0.32\linewidth}
\centering
\qualimg{#1}{0.27\textheight}
\end{minipage}}
\newcommand{\qualpanelwidth}[1]{%
\begin{minipage}[t]{0.32\linewidth}
\centering
\qualimgwidth{#1}
\end{minipage}}
\newcommand{\qualpanelfit}[2]{%
\begin{minipage}[c][#2][c]{0.32\linewidth}
\centering
\qualimgfit{#1}{#2}
\end{minipage}}
\newcommand{\qualpanelcustom}[2]{%
\begin{minipage}[c][#2][c]{0.32\linewidth}
\centering
\qualimg{#1}{#2}
\end{minipage}}
\newcommand{\decodepanel}[5]{%
\begin{minipage}[c][0.76\textheight][c]{0.48\linewidth}
\centering
\textbf{#1}\\[0.5em]
\qualimg{#3}{#2}\\[0.8em]
\qualimg{#5}{#4}
\end{minipage}}

This section presents qualitative examples illustrating the behavior of \mineru{} on representative document pages and diffusion decoding tasks. This section is structured as follows: \Cref{fig:qual_examples_1,fig:qual_examples_2} show complete recognition examples from OmniDocBench, covering the full pipeline from the input page to layout prediction and rendered parsed output. \Cref{fig:decode_layout,fig:decode_text,fig:decode_table,fig:decode_formula} visualize diffusion decoding for layout, text, table, and formula generation.

\subsection{Complete Recognition Examples}

\begin{figure}[p]
\centering
\textbf{Case 1. Academic literature}\\[0.4em]
\qualpanelfit{}{0.255\textheight}\hfill
\qualpanelfit{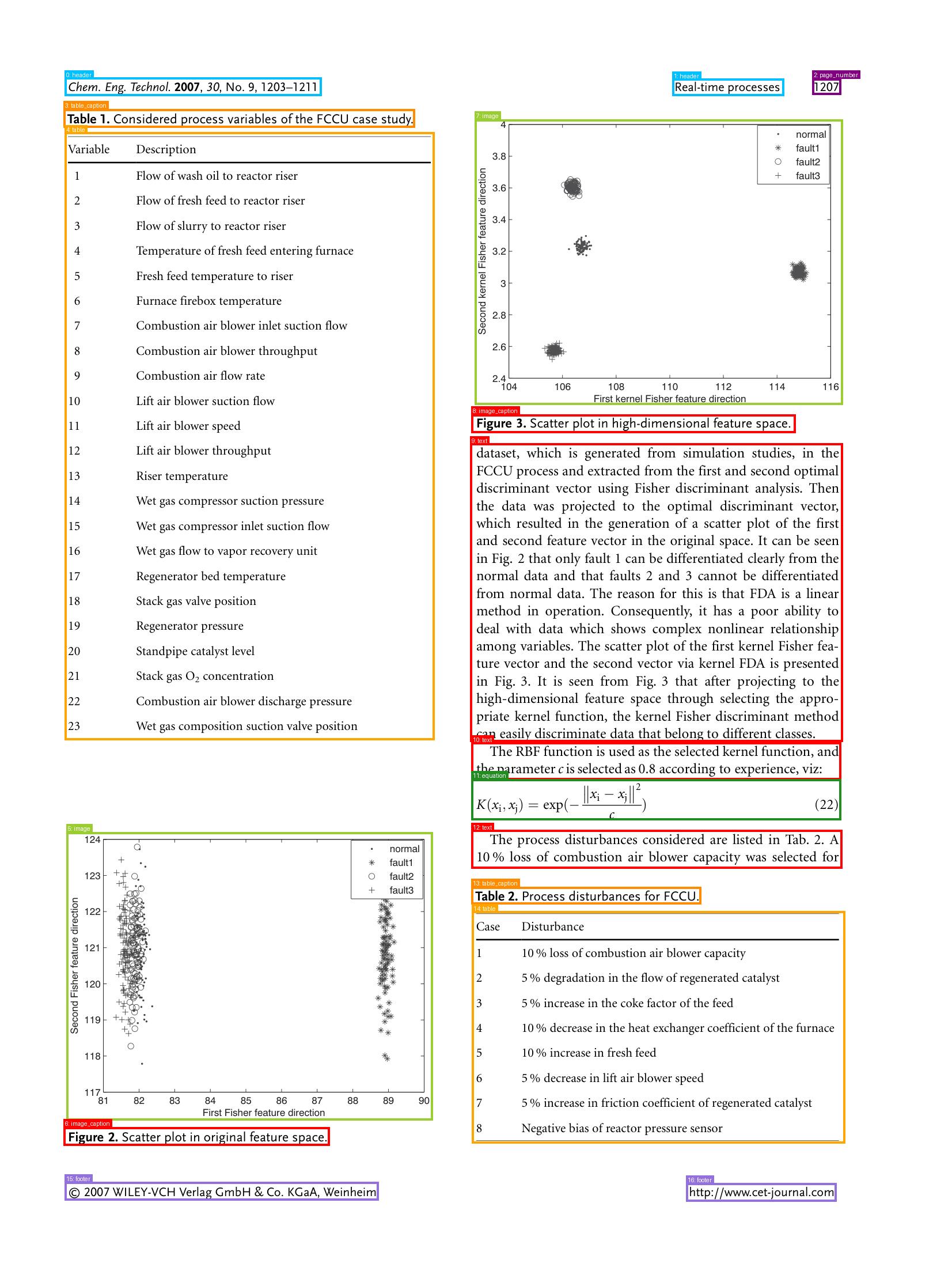}{0.255\textheight}\hfill
\qualpanelfit{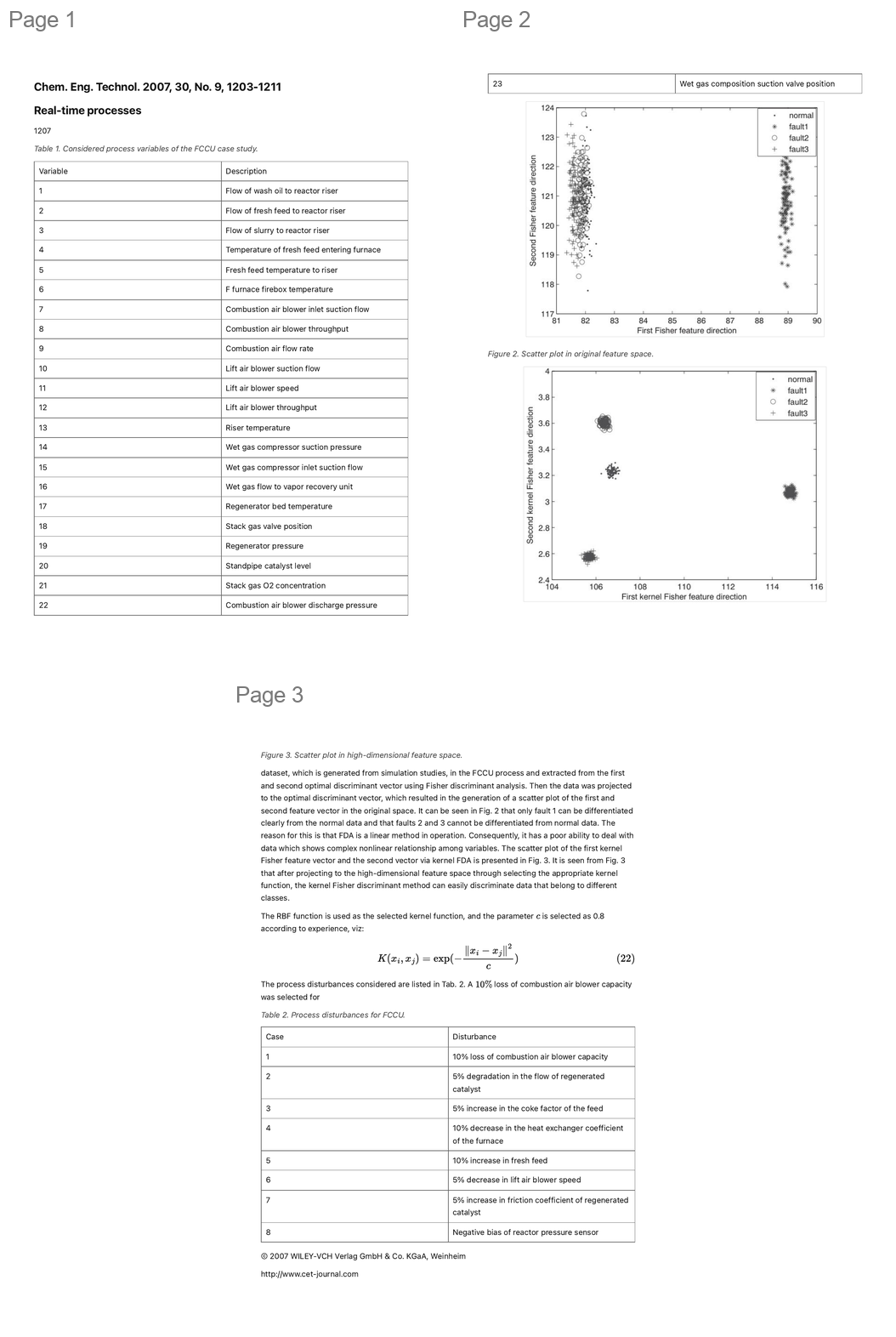}{0.255\textheight}

\textbf{Case 2. Watermarked PPT page}\\[0.4em]
\qualpanelfit{}{0.14\textheight}\hfill
\qualpanelfit{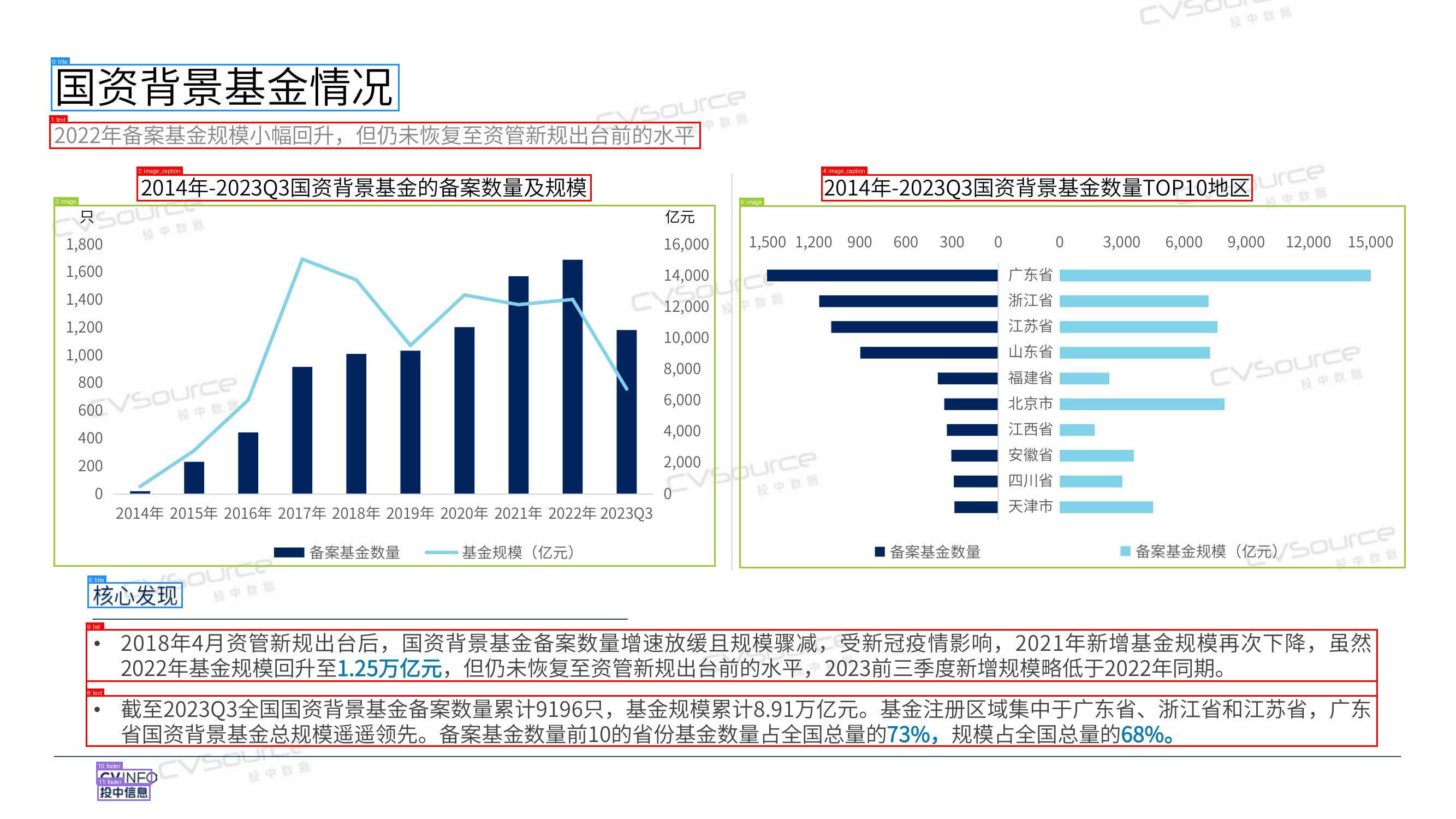}{0.14\textheight}\hfill
\qualpanelfit{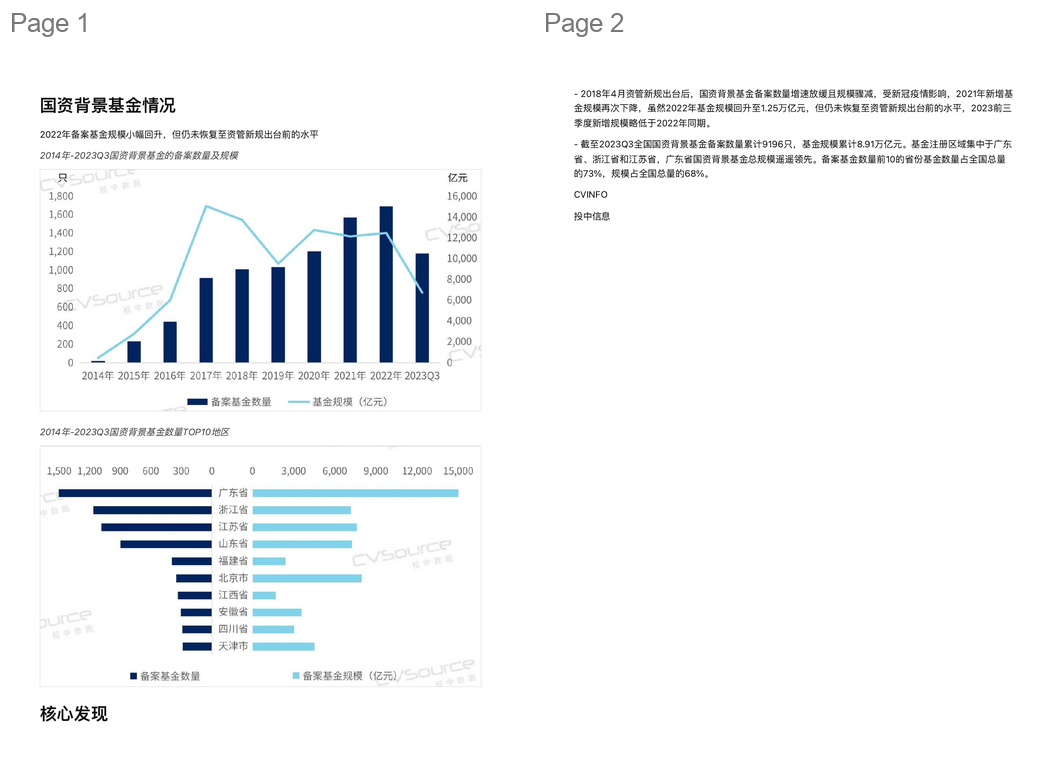}{0.14\textheight}

\textbf{Case 3. Colorful report page}\\[0.4em]
\qualpanelfit{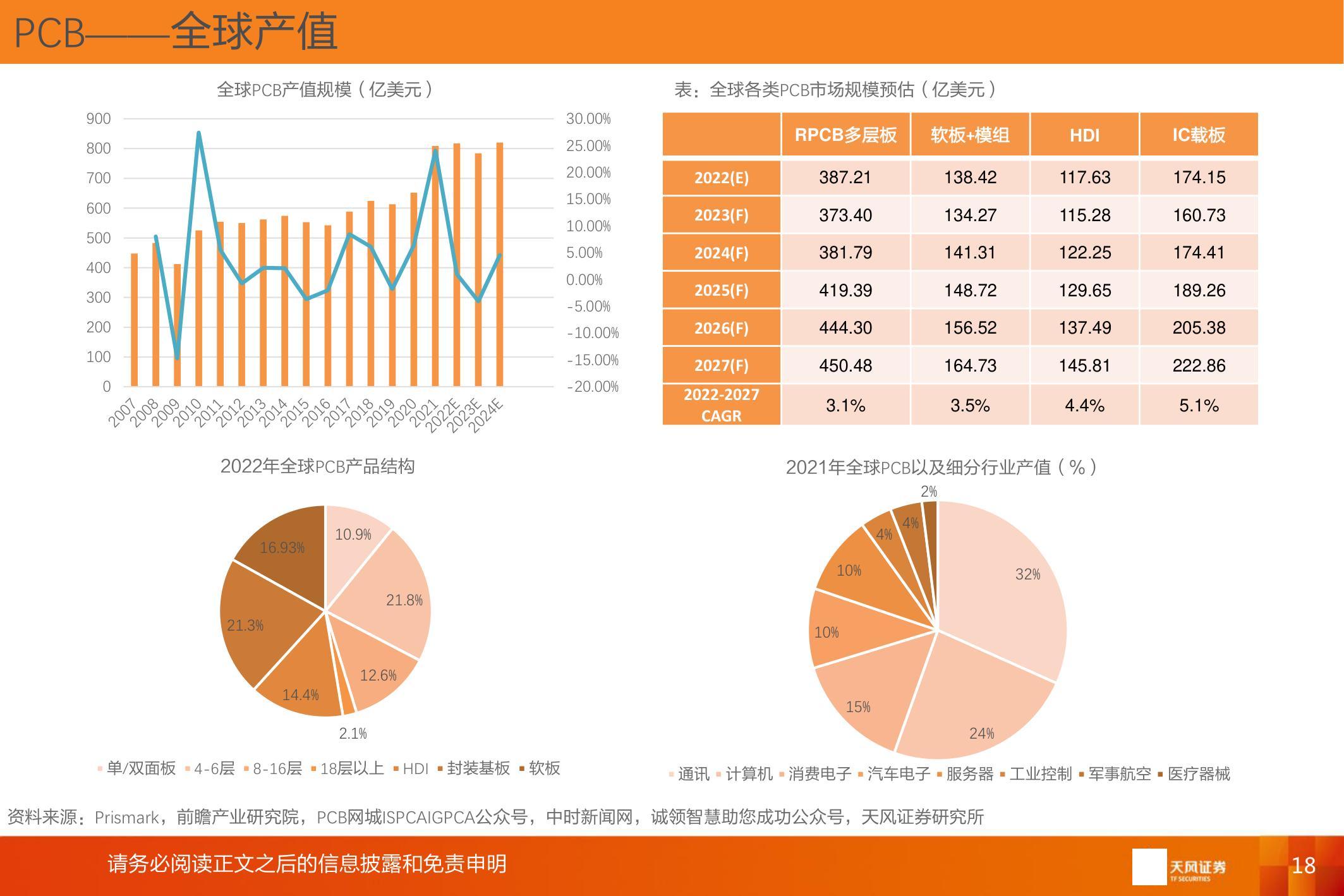}{0.16\textheight}\hfill
\qualpanelfit{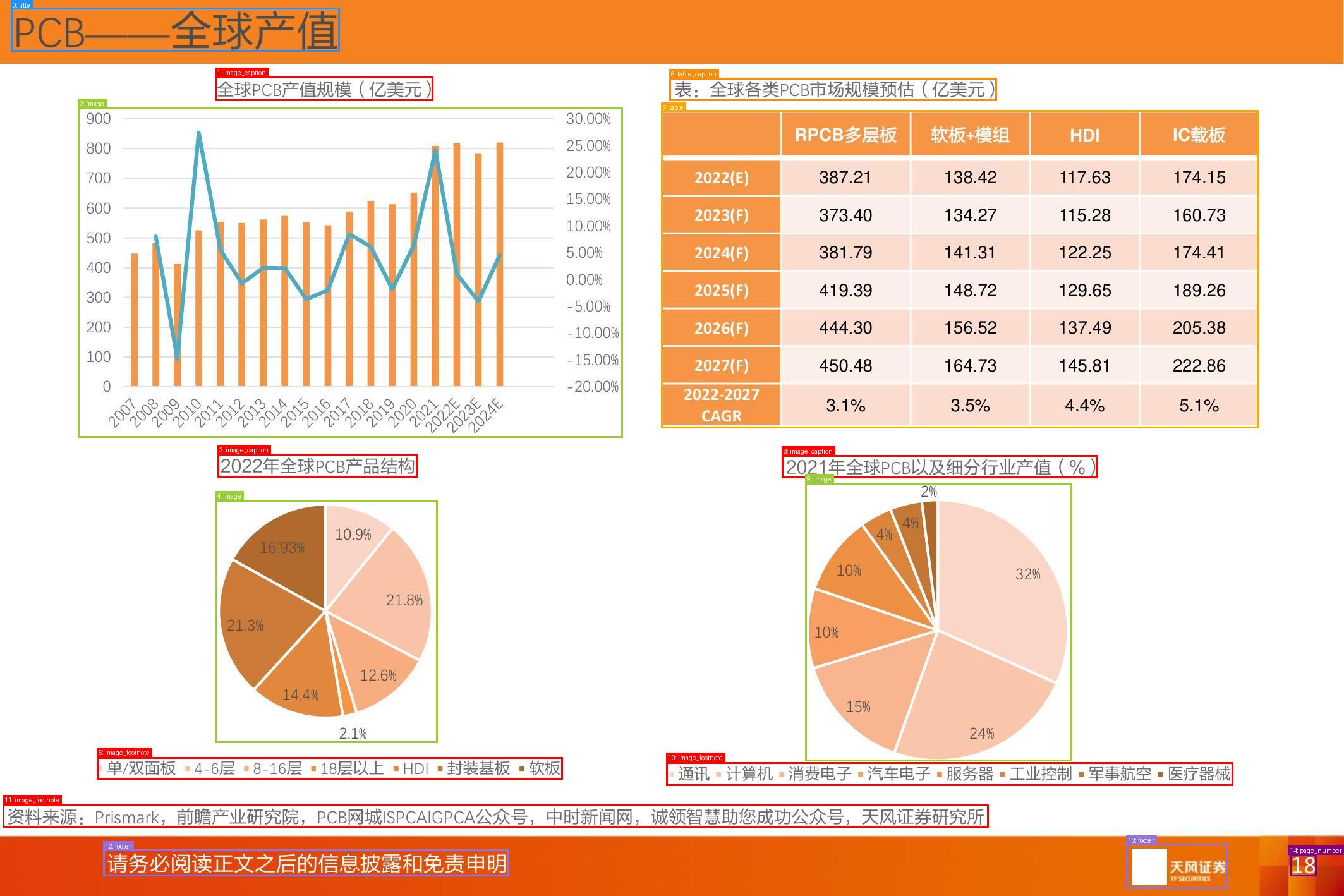}{0.16\textheight}\hfill
\qualpanelfit{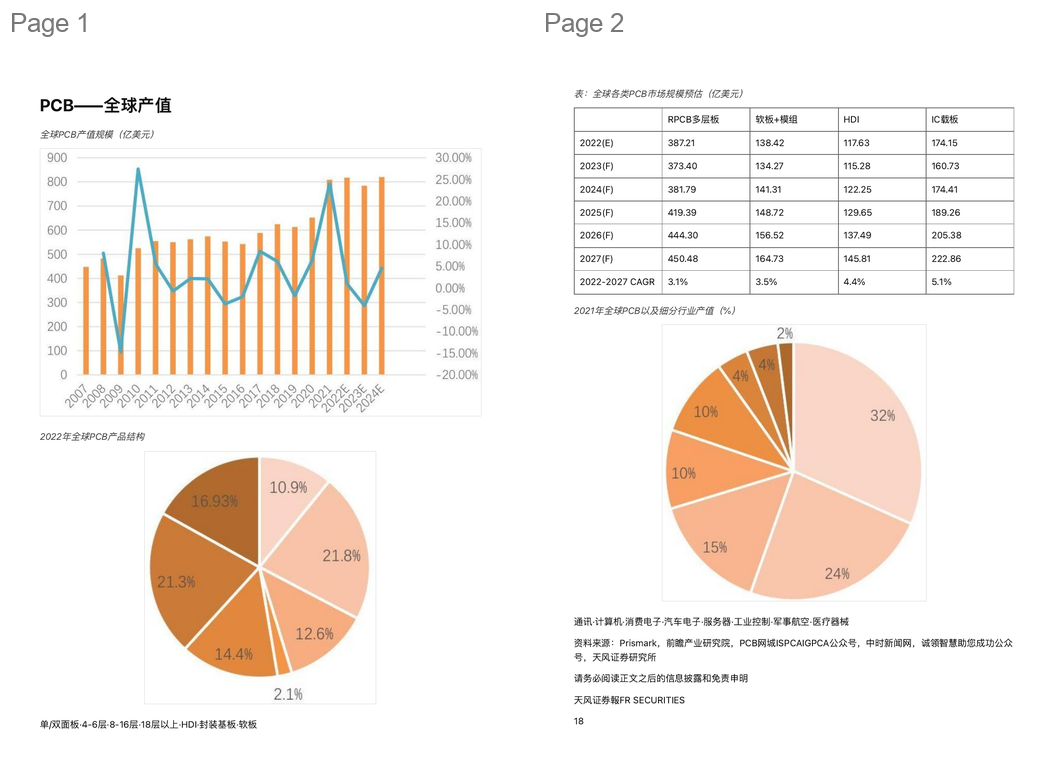}{0.16\textheight}
\caption{Complete recognition examples. Left: original page image. Middle: predicted layout. Right: rendered parsed output.}
\label{fig:qual_examples_1}
\end{figure}

\begin{figure}[p]
\centering
\textbf{Case 4. Newspaper page}\\[0.4em]
\qualpanelfit{}{0.25\textheight}\hfill
\qualpanelfit{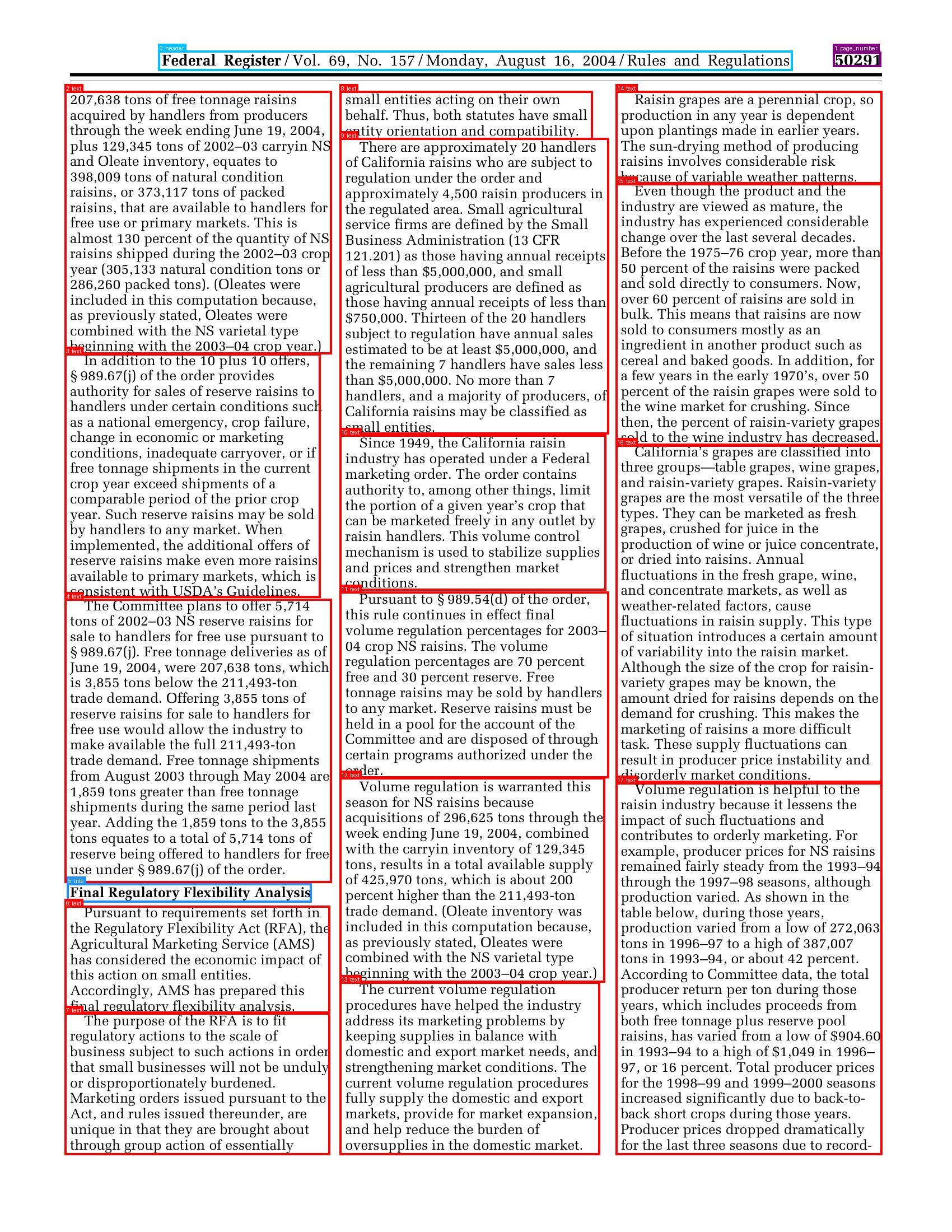}{0.25\textheight}\hfill
\qualpanelfit{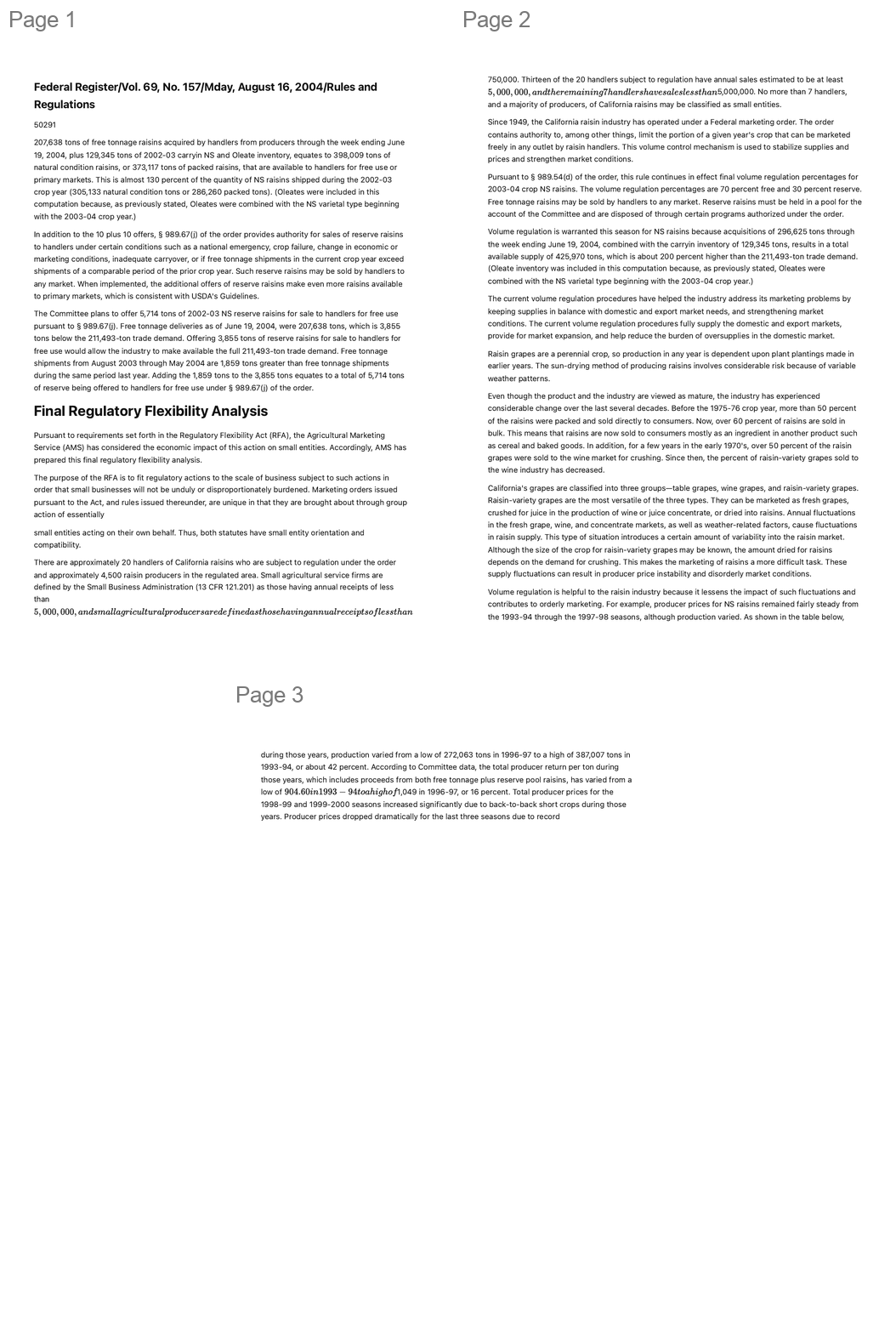}{0.25\textheight}

\textbf{Case 5. Formula-dense exam solution}\\[0.4em]
\qualpanelfit{}{0.235\textheight}\hfill
\qualpanelfit{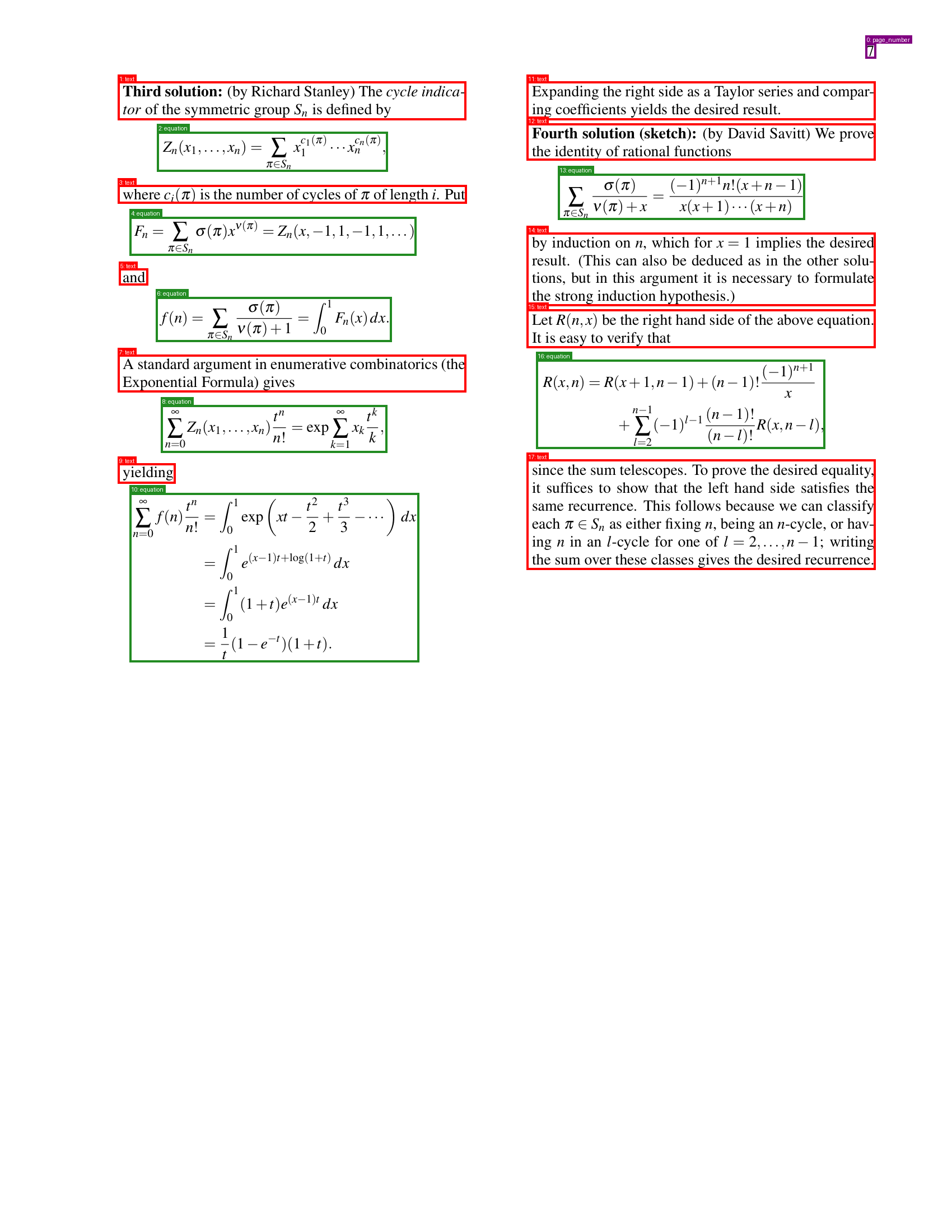}{0.235\textheight}\hfill
\qualpanelfit{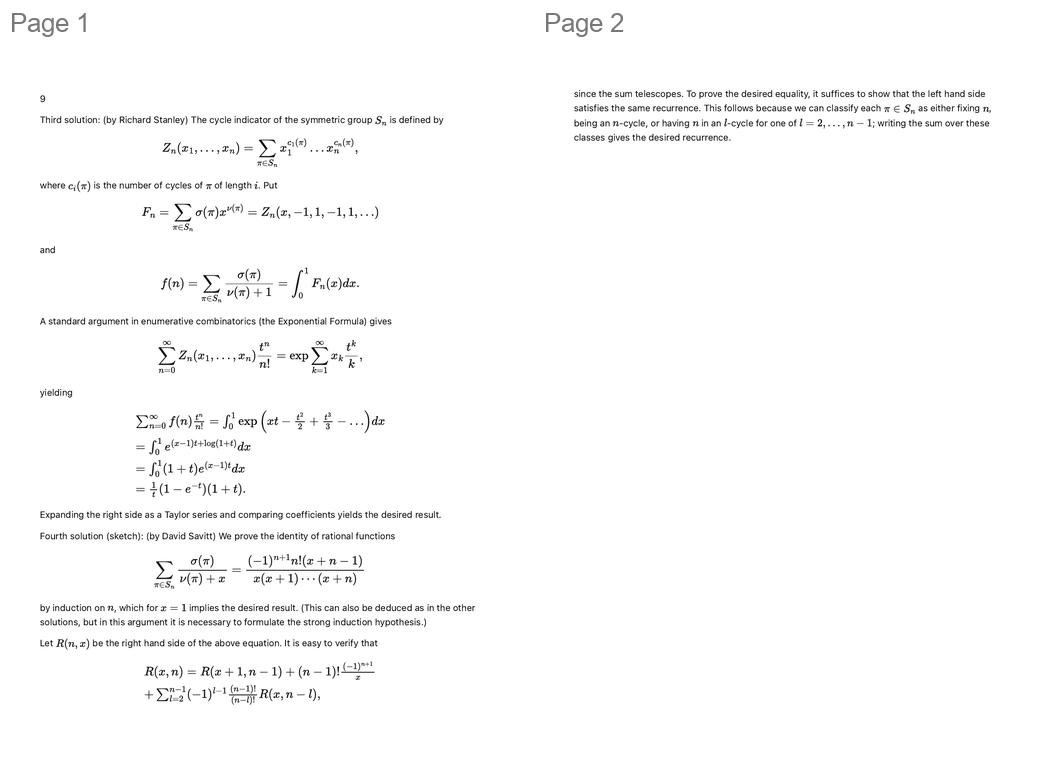}{0.235\textheight}

\textbf{Case 6. Illustrated recipe page}\\[0.4em]
\qualpanelfit{}{0.235\textheight}\hfill
\qualpanelfit{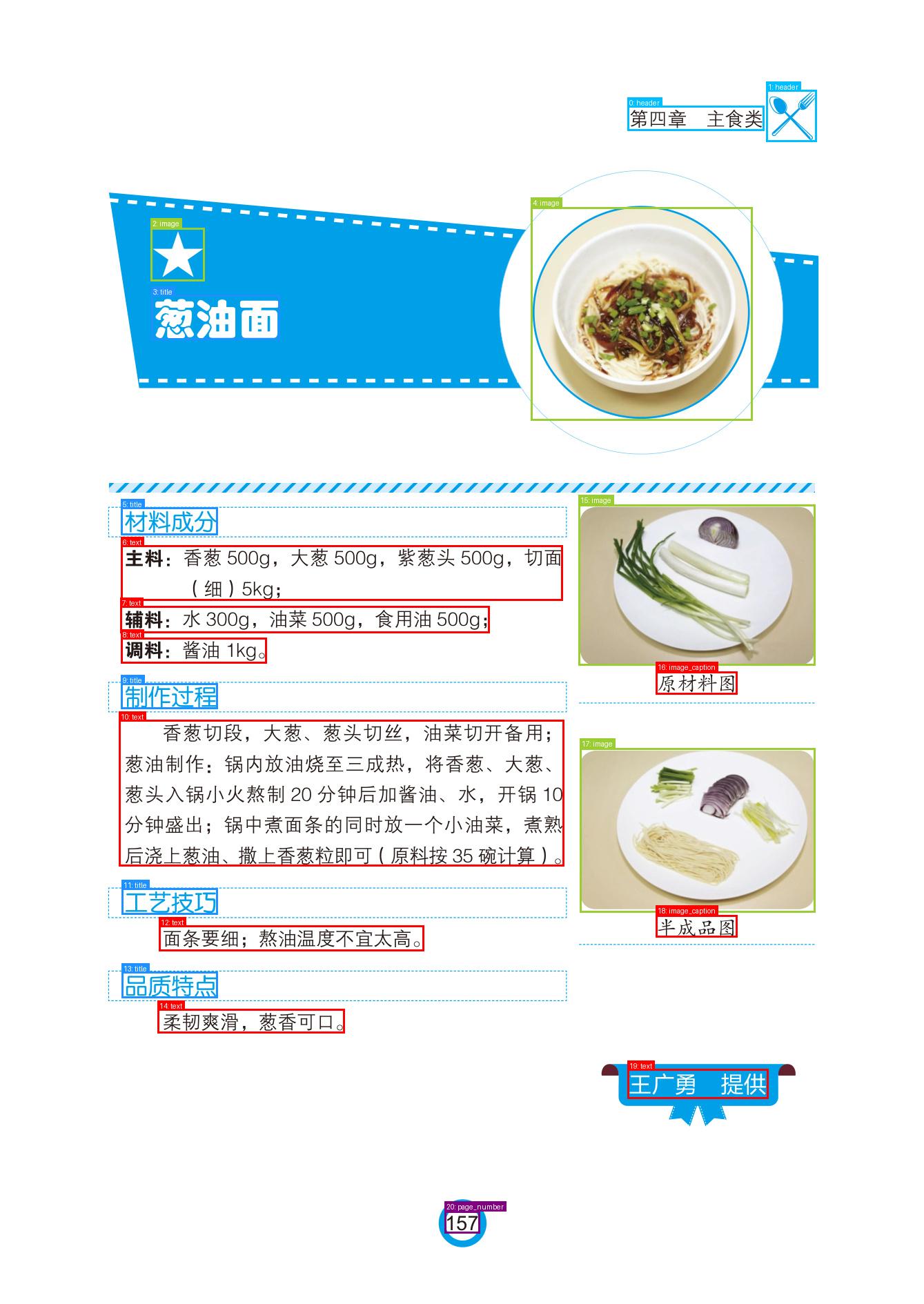}{0.235\textheight}\hfill
\qualpanelfit{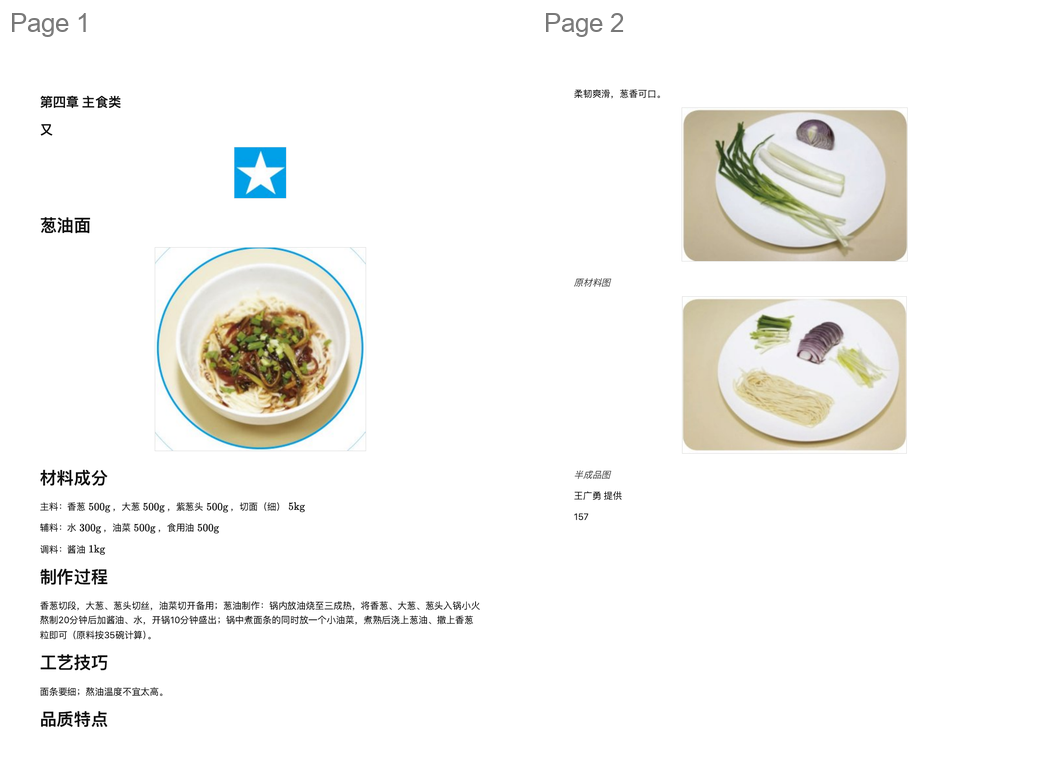}{0.235\textheight}
\caption{Additional complete recognition examples in the same format.}
\label{fig:qual_examples_2}
\end{figure}

Examples demonstrating end-to-end parsing performance on diverse page types are provided in \Cref{fig:qual_examples_1,fig:qual_examples_2}, including academic literature, presentation slides, colorful report pages, newspapers, formula-dense exam solutions, and illustrated instructional pages. These examples show that \mineru{} can preserve complex academic elements such as tables, figures, captions, and equations, remain robust to watermark interference, maintain reading order on dense multi-column pages, and recover mixed text-image layouts with fine-grained blocks. Each case is shown in a consistent three-column format with the original page image, the predicted layout, and the rendered parsed output.

\subsection{Diffusion Decoding Examples}

\begin{figure}[p]
\centering
\textbf{Layout Decoding}\\[0.8em]
\decodepanel{Example 1}{0.18\textheight}{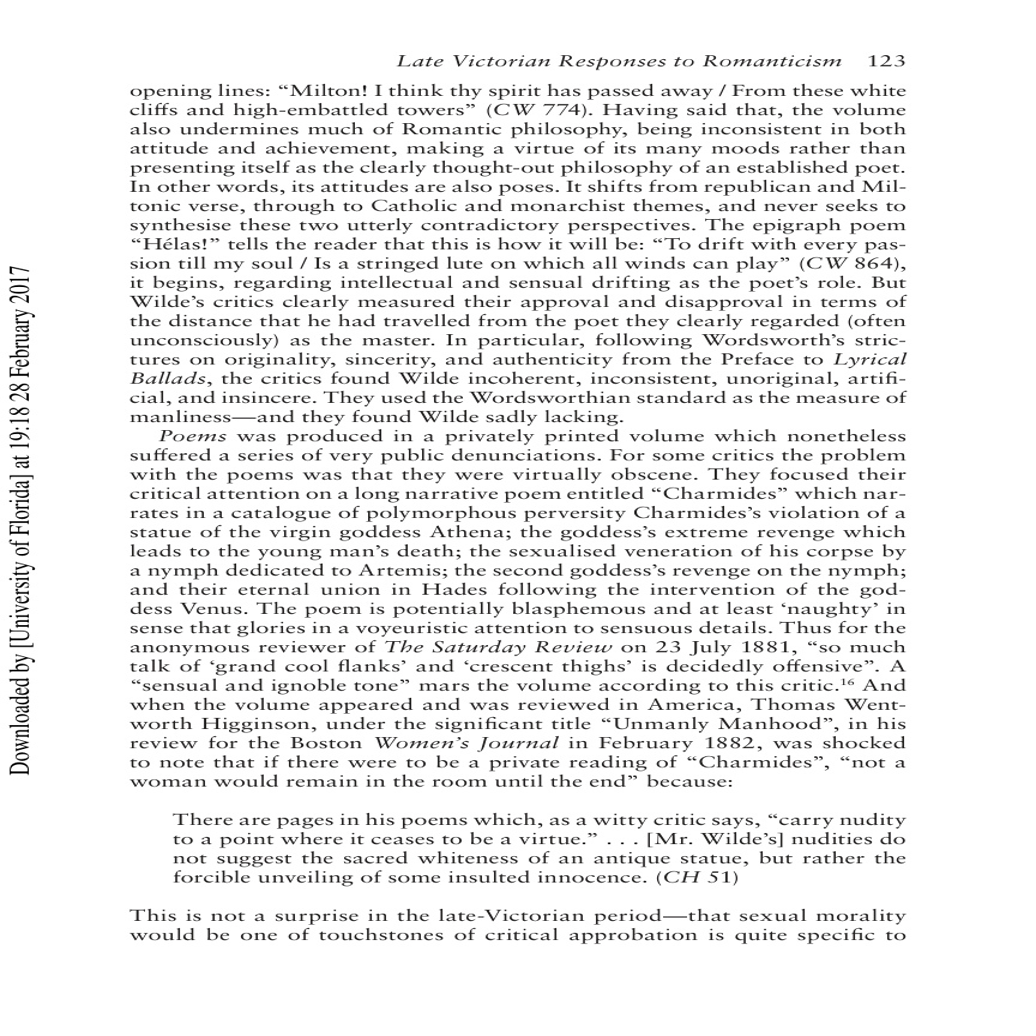}{0.56\textheight}{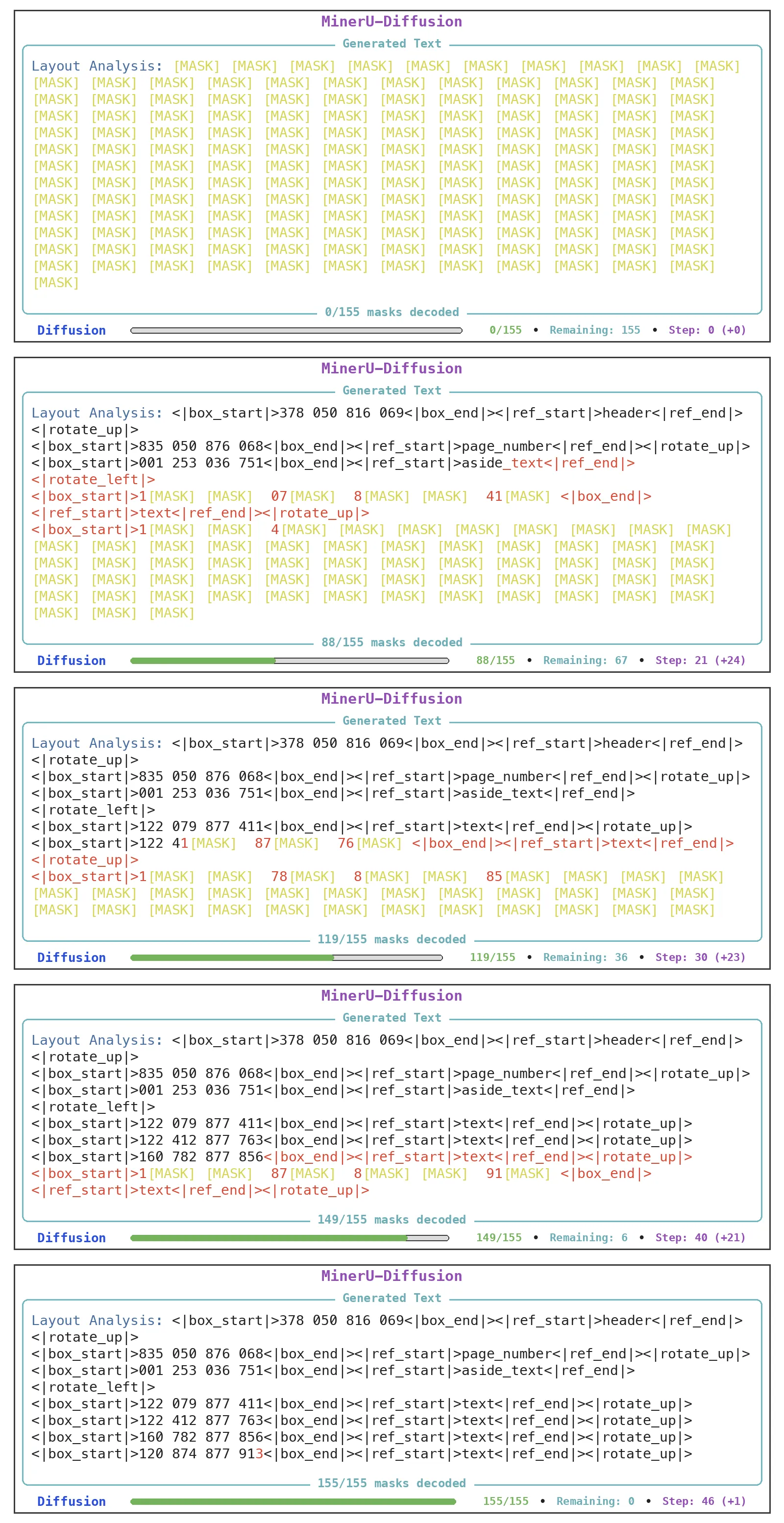}\hfill
\decodepanel{Example 2}{0.18\textheight}{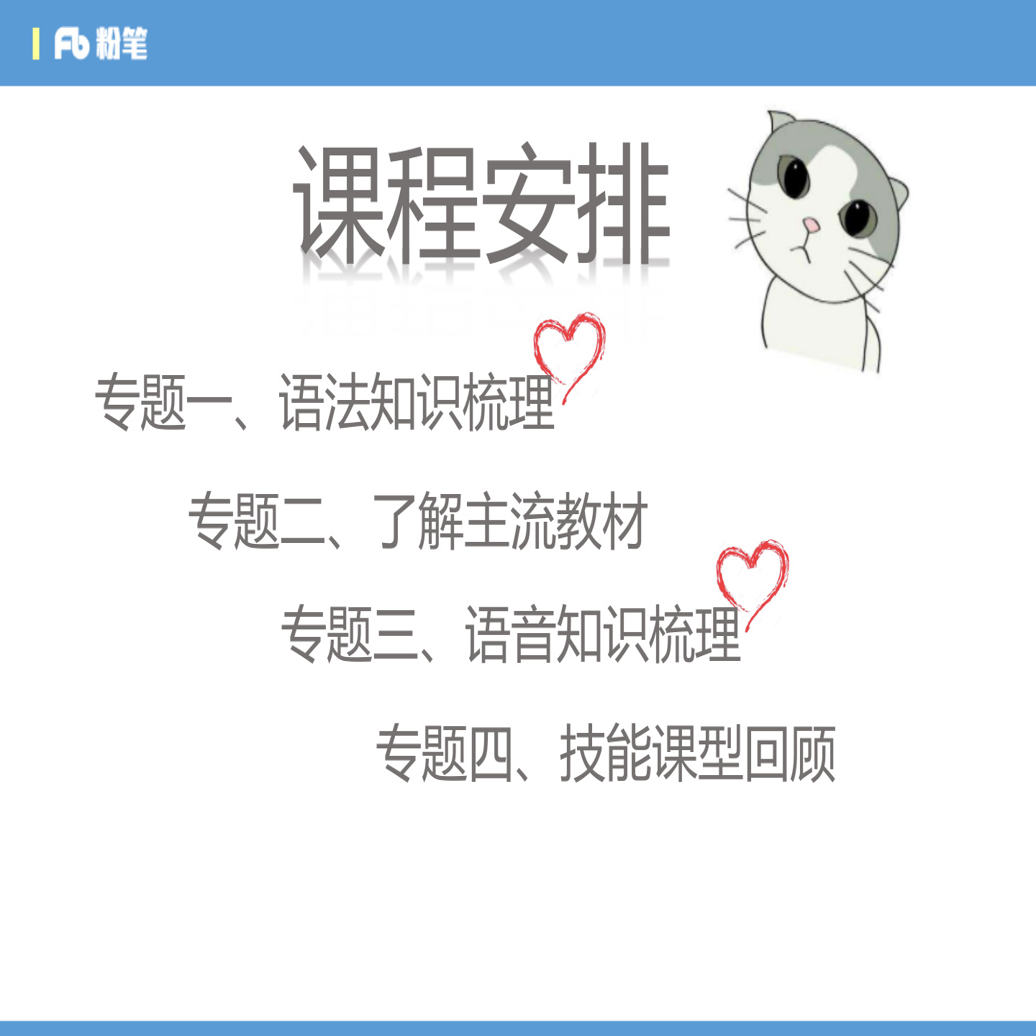}{0.56\textheight}{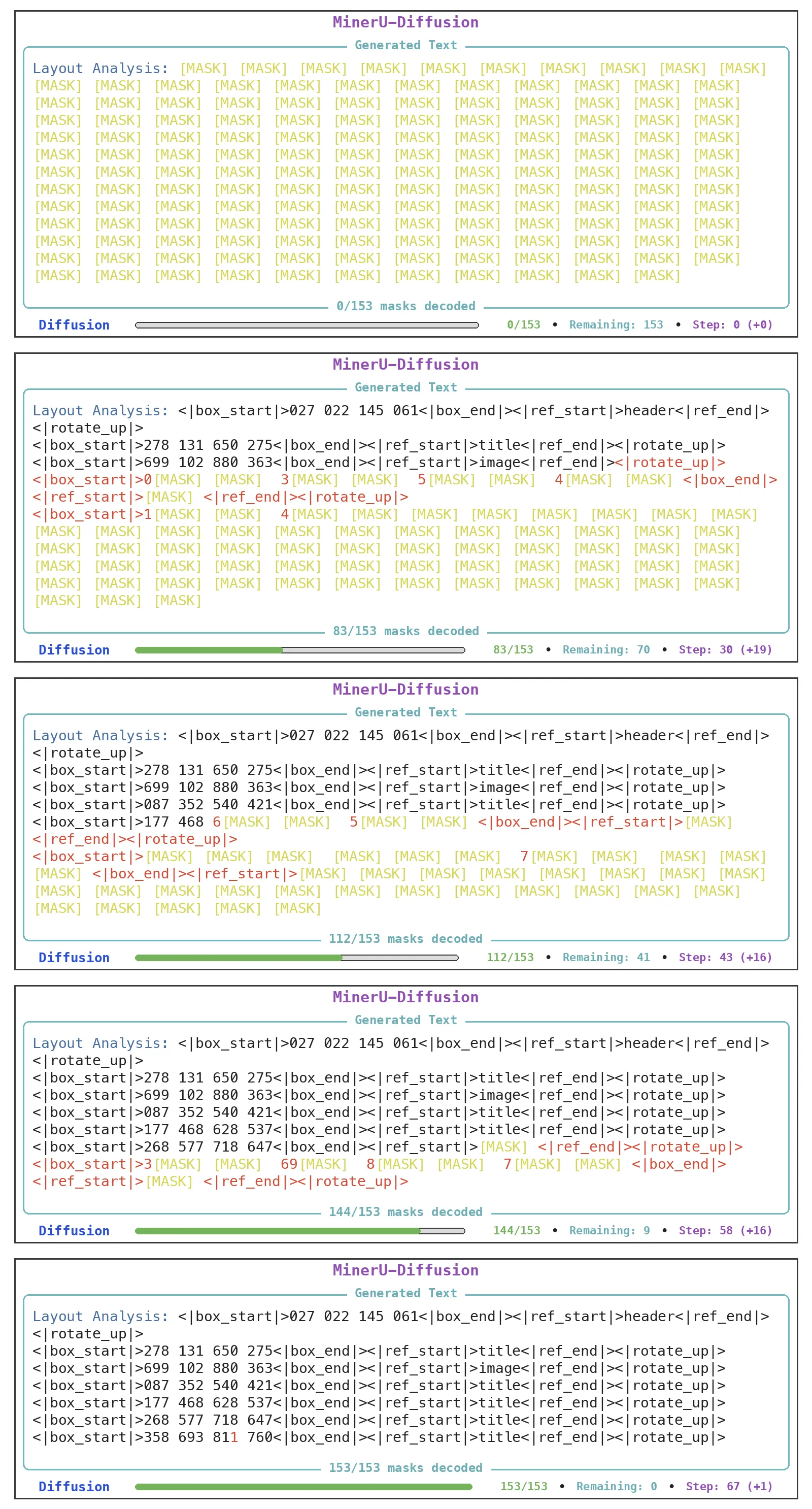}
\caption{Diffusion decoding examples for layout generation.}
\label{fig:decode_layout}
\end{figure}

\begin{figure}[p]
\centering
\textbf{Text Decoding}\\[0.8em]
\decodepanel{Example 1}{0.08\textheight}{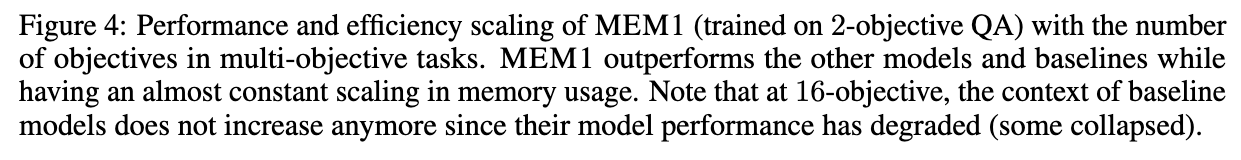}{0.58\textheight}{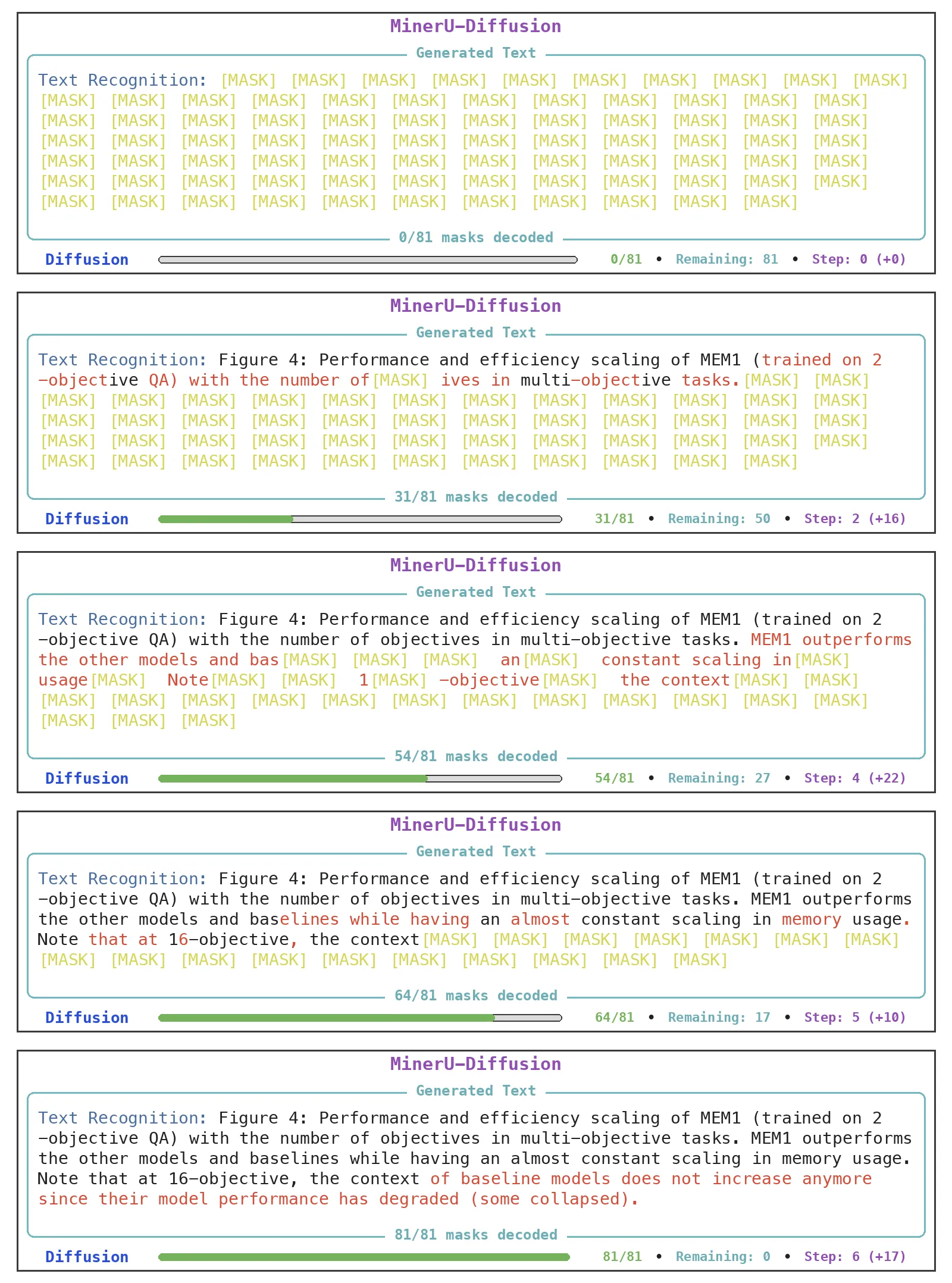}\hfill
\decodepanel{Example 2}{0.08\textheight}{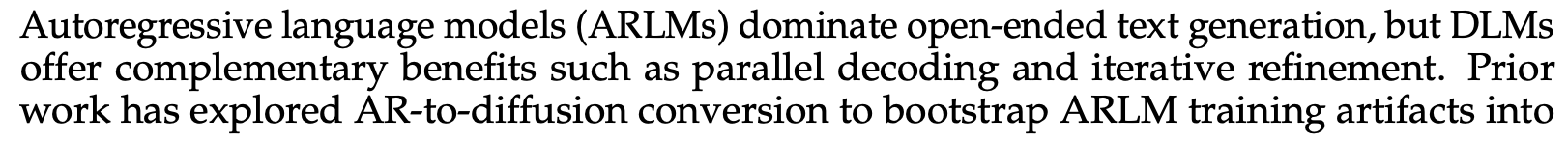}{0.58\textheight}{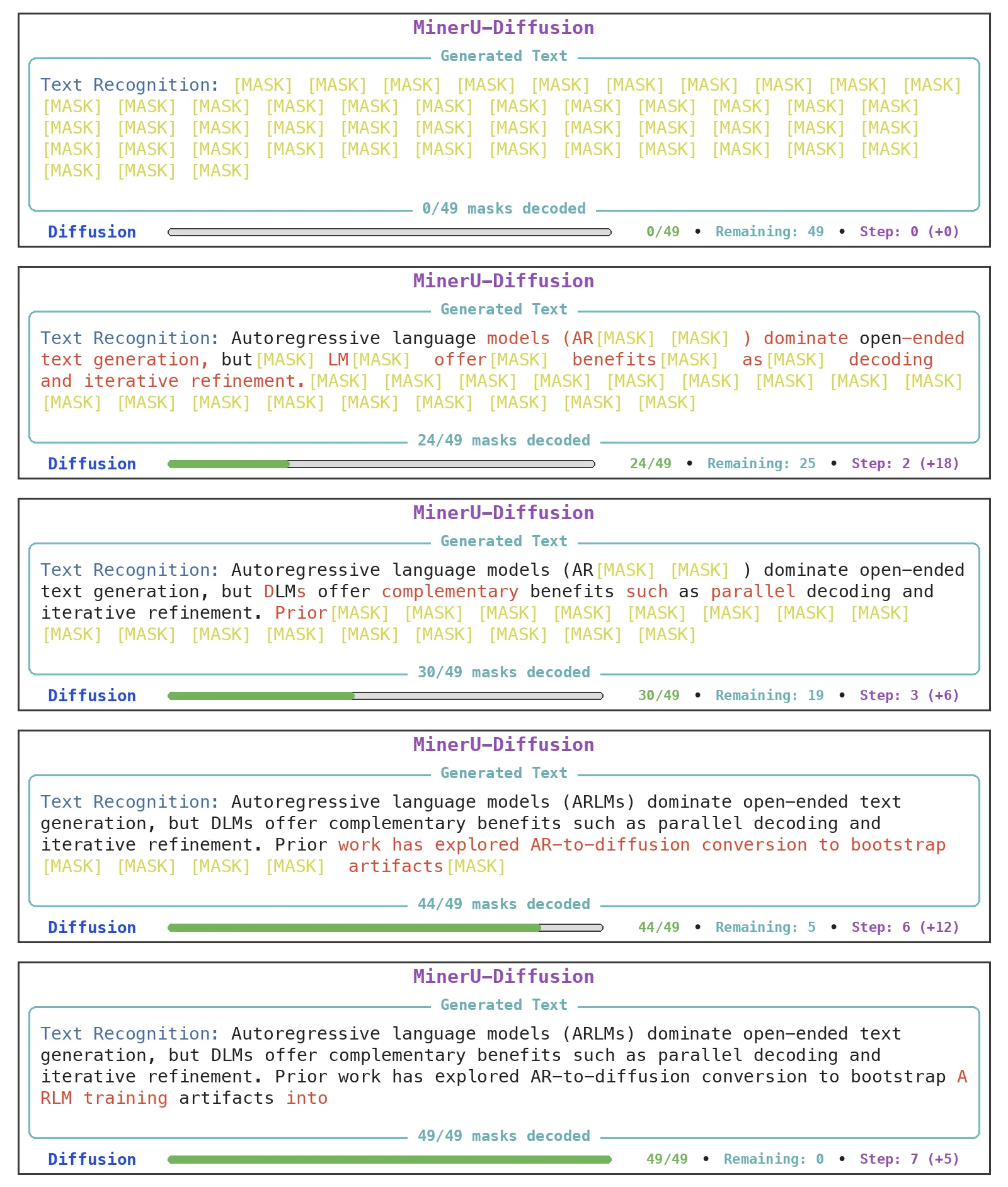}
\caption{Diffusion decoding examples for text recognition.}
\label{fig:decode_text}
\end{figure}

\begin{figure}[p]
\centering
\textbf{Table Decoding}\\[0.8em]
\decodepanel{Example 1}{0.11\textheight}{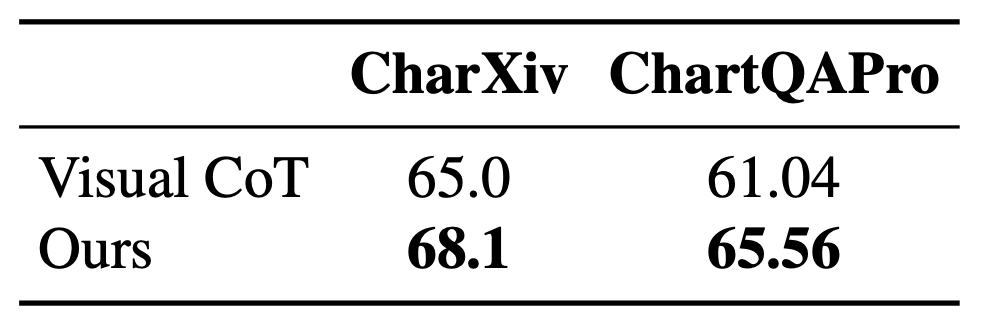}{0.55\textheight}{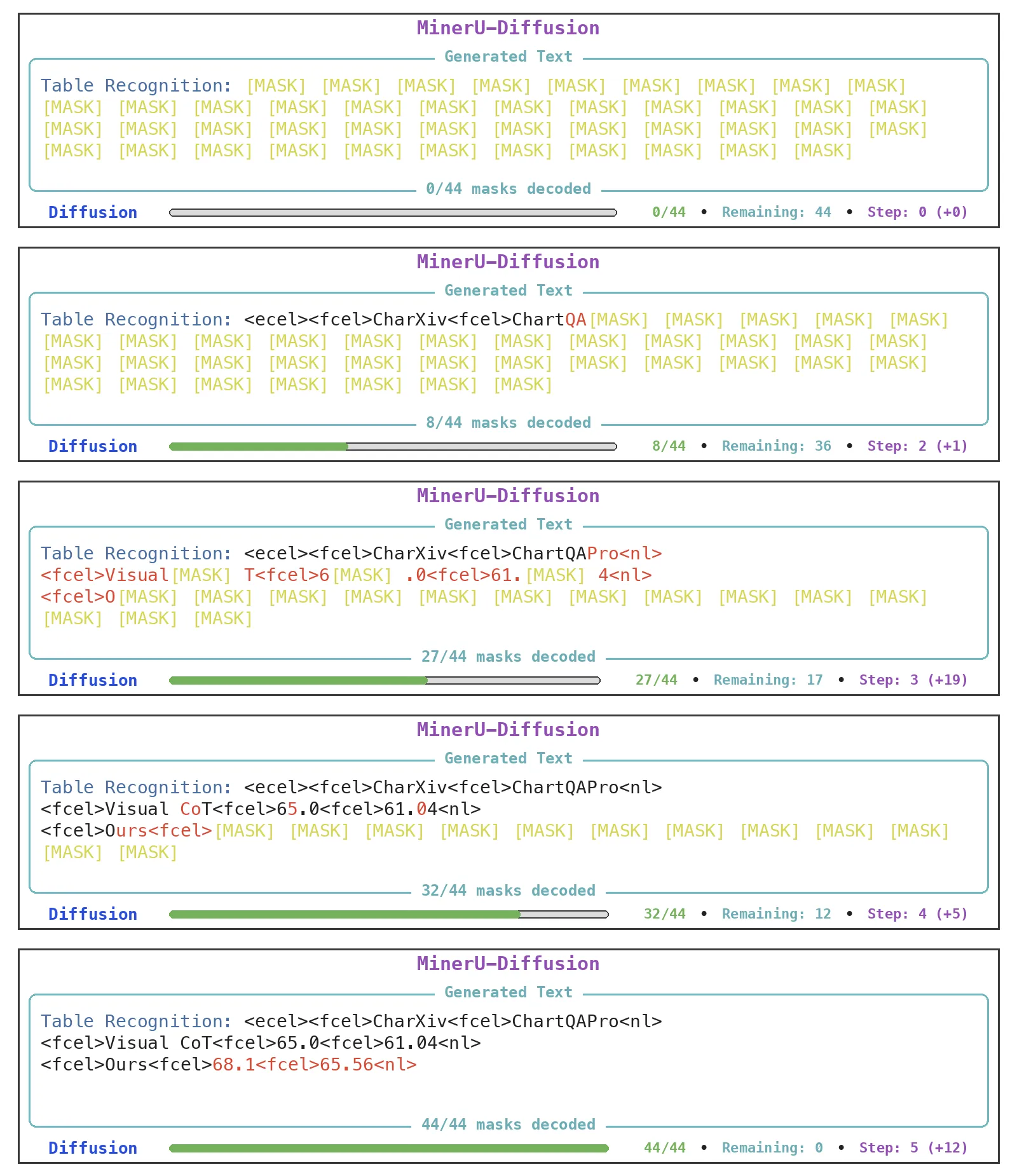}\hfill
\decodepanel{Example 2}{0.11\textheight}{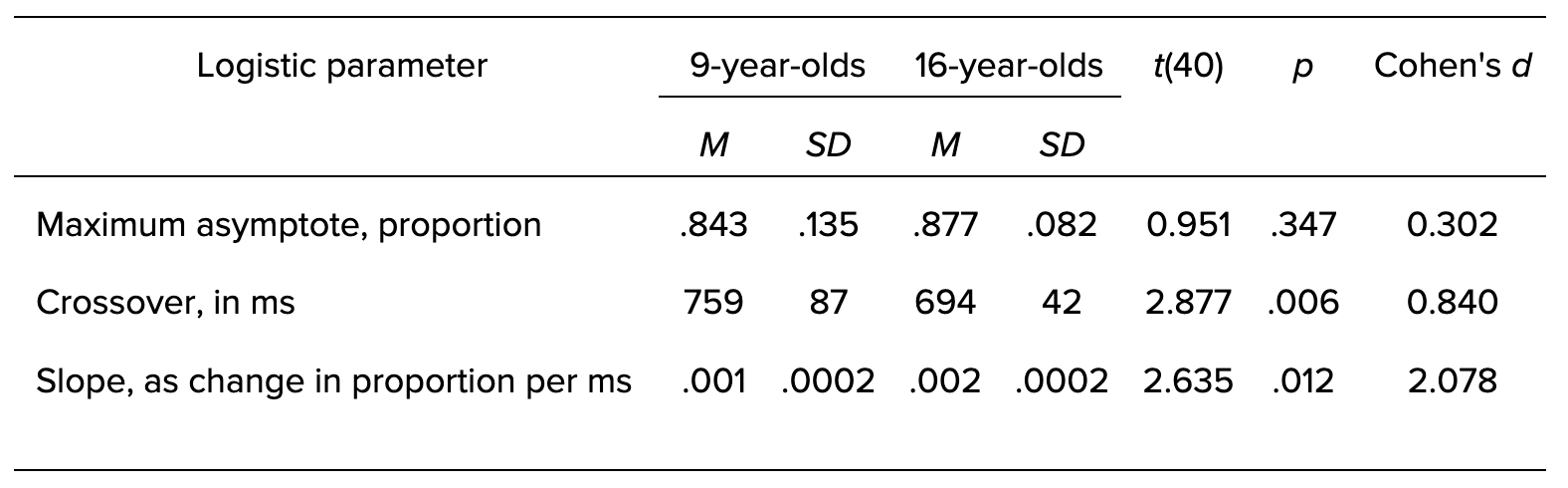}{0.55\textheight}{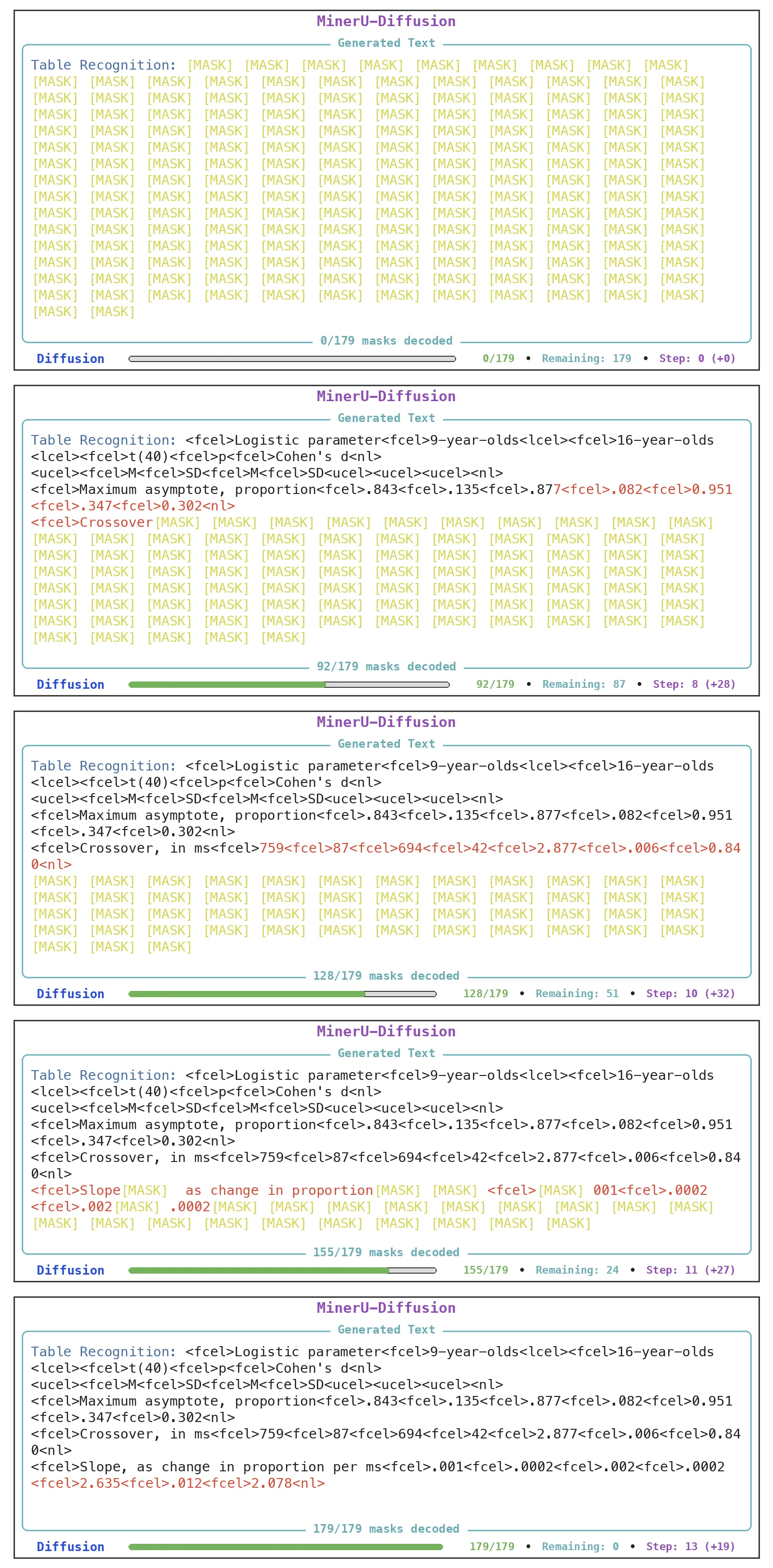}
\caption{Diffusion decoding examples for table recognition.}
\label{fig:decode_table}
\end{figure}

\begin{figure}[p]
\centering
\textbf{Formula Decoding}\\[0.8em]
\decodepanel{Example 1}{0.08\textheight}{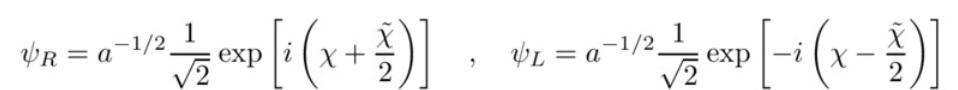}{0.58\textheight}{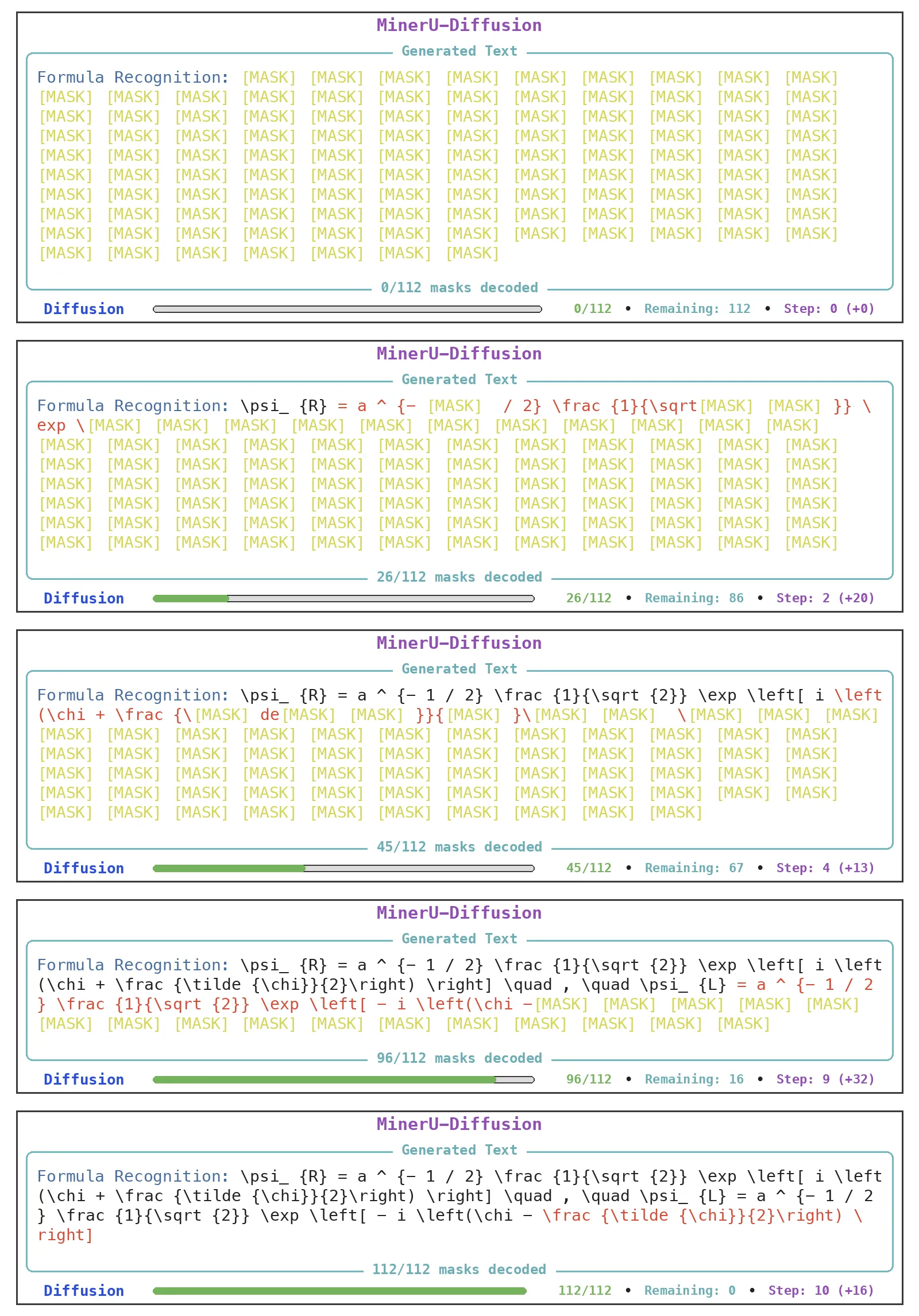}\hfill
\decodepanel{Example 2}{0.08\textheight}{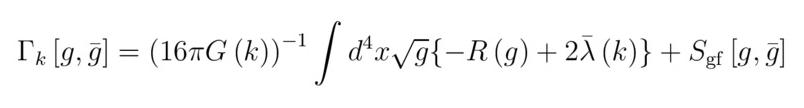}{0.58\textheight}{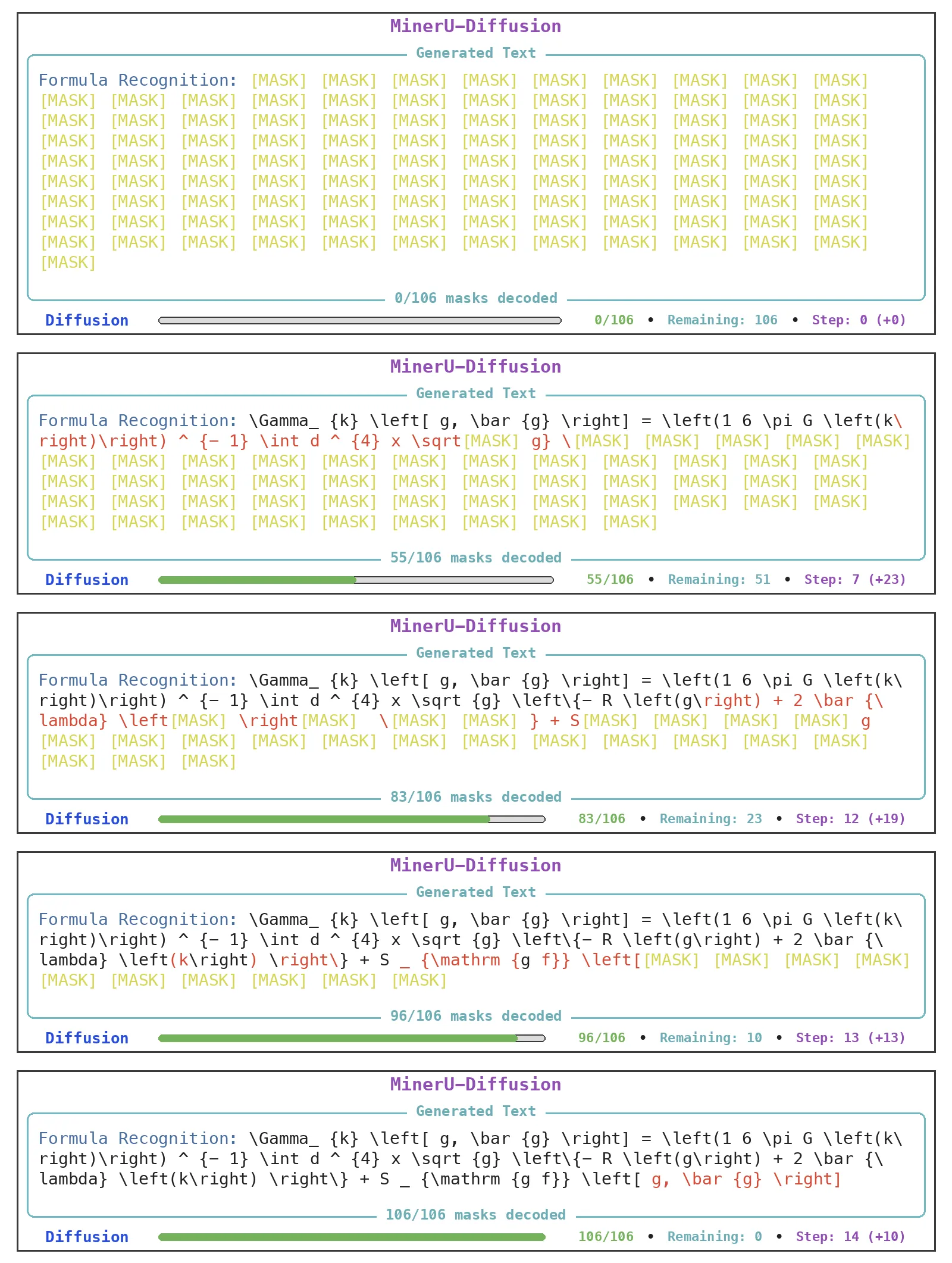}
\caption{Diffusion decoding examples for formula recognition.}
\label{fig:decode_formula}
\end{figure}

The diffusion decoding process on four representative tasks is illustrated in \Cref{fig:decode_layout,fig:decode_text,fig:decode_table,fig:decode_formula}. These examples show how the model progressively refines page structure, text content, table organization, and formula symbols during decoding. Each figure groups two samples from the same category to make the visual patterns directly comparable, with the raw cropped input shown above and the corresponding diffusion decoding visualization shown below.

\clearpage

\clearpage

\clearpage

\clearpage

\clearpage

\clearpage

\end{document}